\documentclass[sigconf,nonacm]{acmart}
\AtBeginDocument{%
  }

\usepackage{microtype}
\usepackage{graphicx}
\usepackage{subcaption}
\usepackage{booktabs} 

\usepackage{hyperref}
\usepackage{comment}
\usepackage{placeins} 
\usepackage{color}
\usepackage{soul}
\usepackage{url}
\usepackage[utf8]{inputenc}
\usepackage{algpseudocode}
\usepackage[switch]{lineno}
\usepackage{multirow}
\usepackage{graphicx}
\usepackage{dblfloatfix}
\usepackage[table]{xcolor} 
\usepackage{amsmath}

\usepackage{algorithm}

\usepackage{mathtools}
\usepackage{amsthm}
\usepackage{booktabs}

\begin{document}

\title{VideoMM: Adaptive Macro-Micro Inference for Efficient Video MLLMs}

\author{Haoyu Guo\textsuperscript{1,3,$\ast$}{\char44} Yuan Feng\textsuperscript{2,3,$\ast$}{\char44} Junlin Lv\textsuperscript{2,3}{\char44} Mingjun Xiao\textsuperscript{2,3}{\char44} S Kevin Zhou\textsuperscript{1,3}{\char44} Xike Xie\textsuperscript{1,3,$\dagger$}}

\renewcommand{\shortauthors}{Guo et al.}

\affiliation{%
  \institution{
    \textsuperscript{1}School of Biomedical Engineering, University of Science and Technology of China \\
    \textsuperscript{2}School of Computer Science, USTC \\
    \textsuperscript{3}Data Darkness Lab, MIRACLE Center, Suzhou Institute for Advanced Research
  }
  \country{China}
}

\thanks{\textsuperscript{$\ast$}Equal Contribution: \{haoyuguo, yfung\}@mail.ustc.edu.cn}
\thanks{\textsuperscript{$\dagger$}Corresponding author: xkxie@ustc.edu.cn}

\begin{abstract}
Scaling Multimodal Large Language Models (MLLMs) to long-form video understanding is bottlenecked by the explosion of visual tokens, which saturates context windows and incurs prohibitive costs. Current solutions predominantly rely on auxiliary models for token reduction but face a fundamental dilemma: lightweight encoder-driven approaches often overlook critical semantic information, whereas heavyweight MLLM-driven reduction negates the efficiency gains. 
{In this work, we identify a more fundamental inefficiency underlying this dilemma:  while fine-grained visual details are essential for detailed understanding, they are largely  redundant for the preliminary task of selecting semantically relevant regions.
} Motivated by this, we introduce \textbf{VideoMM},  which marks a paradigm shift from model-centric downsizing to adaptive perceptual granularity. Specifically, our framework {decouples selection from reasoning} by executing semantic filtering on a cost-effective \textit{Macro Proxy} (derived from downscaled frames), and projecting the selected regions onto high-fidelity \textit{Micro Tokens} for detailed understanding only when necessary.
Extensive evaluations show that VideoMM significantly outperforms existing solutions. It achieves a 6.13$\times$ speedup and a 7.4\% accuracy gain over full-context baselines on LongVideoBench, and further accelerates inference by 2.73$\times$ over current leading  methods, establishing a highly scalable paradigm for long-video understanding. Our code is available at: https://github.com/adfh917k/VideoMM.
\end{abstract}

\maketitle

\section{Introduction}\label{sec:introduction}

Multimodal Large Language Models (MLLMs) \cite{qwen3technicalreport, Qwen2.5-VL, glm4v, zohar2024apolloexplorationvideounderstanding, 10982110} have recently achieved remarkable success in interpreting static images and short video clips. However, extending these capabilities to long-form video understanding remains a critical challenge. While long-video analysis enables high-value applications—such as video QA, temporal event localization, and long-horizon action recognition \cite{Liu_2025_CVPR, AlShami_2024, Liu_2021_CVPR}—it introduces severe scalability bottlenecks. The primary constraint is the massive influx of visual tokens: for instance, a single 10-minute 720p video sampled at 1 fps generates over 300k tokens in the widely adopted Qwen-3-VL model. This volume rapidly exhausts the limited context windows of current models, leading to prohibitive computational costs.

\begin{figure}[t]
	\centering
    \includegraphics[width=0.9\linewidth]{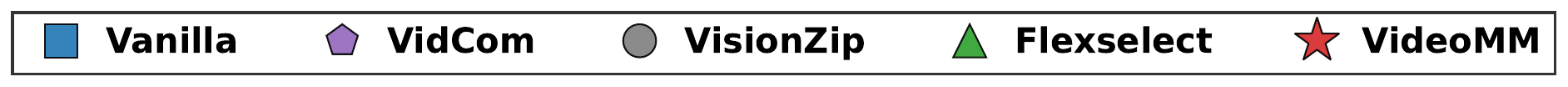}
	\begin{subfigure}{0.33\linewidth}
		\centering
		\includegraphics[width=\linewidth]{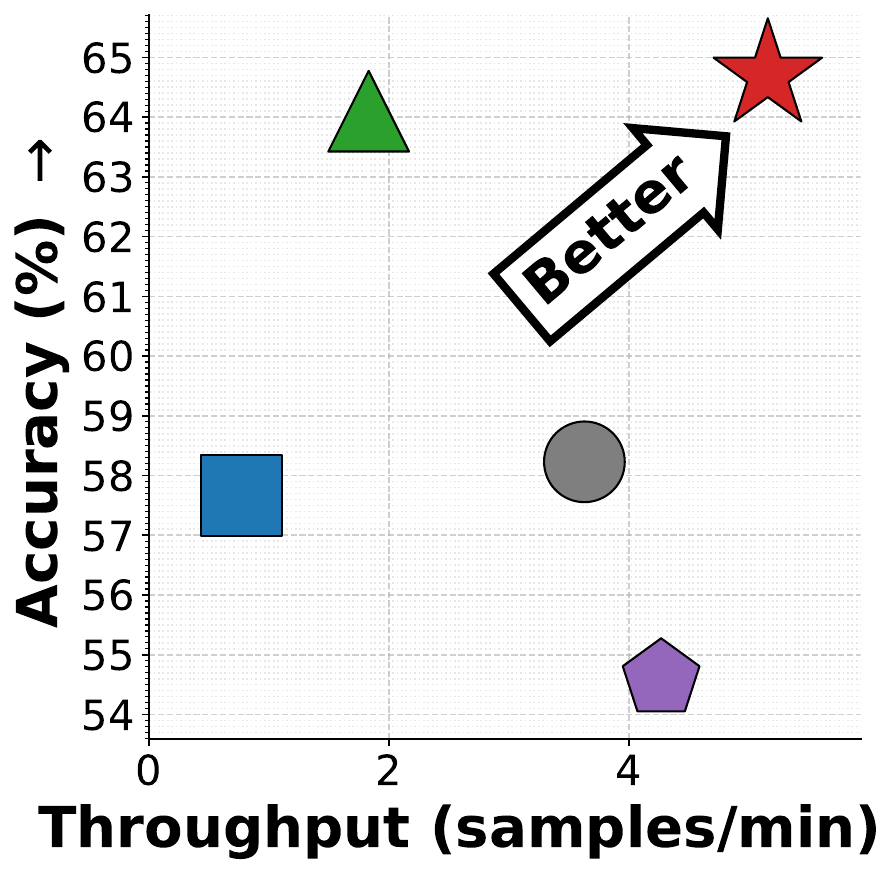}
		\vspace{-0.3cm}
		\caption{Qwen2.5-VL-7B}
	\end{subfigure}%
    \begin{subfigure}{0.33\linewidth}
		\centering
		\includegraphics[width=\linewidth]{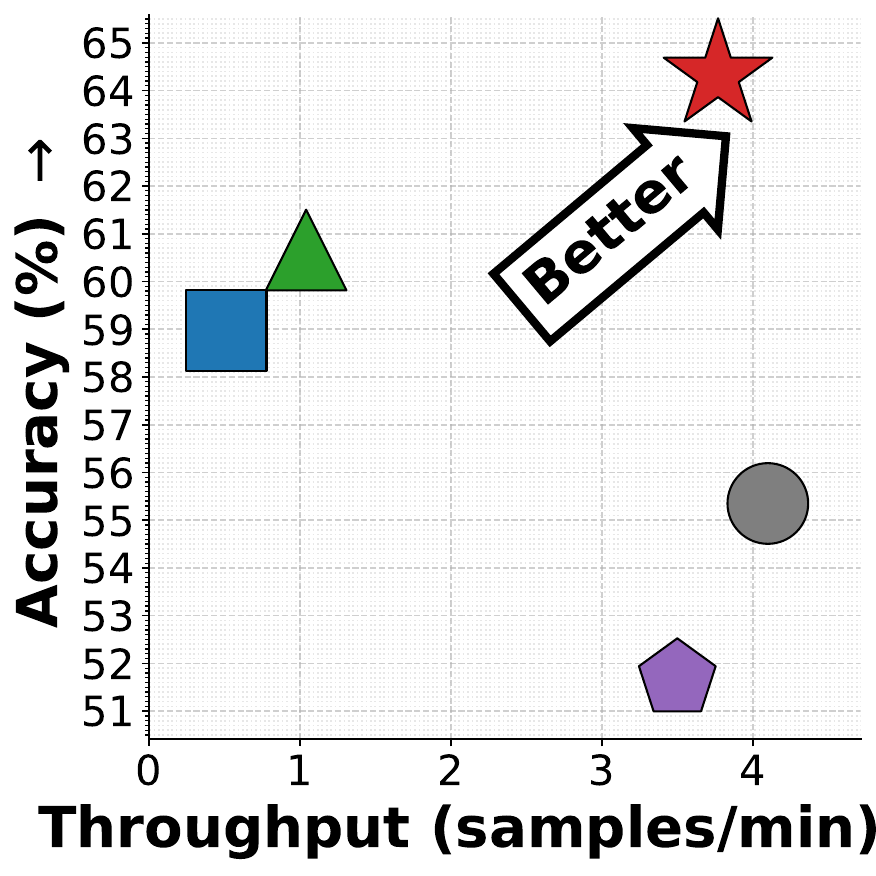}
		\vspace{-0.3cm}
		\caption{GLM-4.1V-9B}
	\end{subfigure}
	\begin{subfigure}{0.33\linewidth}
		\centering
		\includegraphics[width=\linewidth]{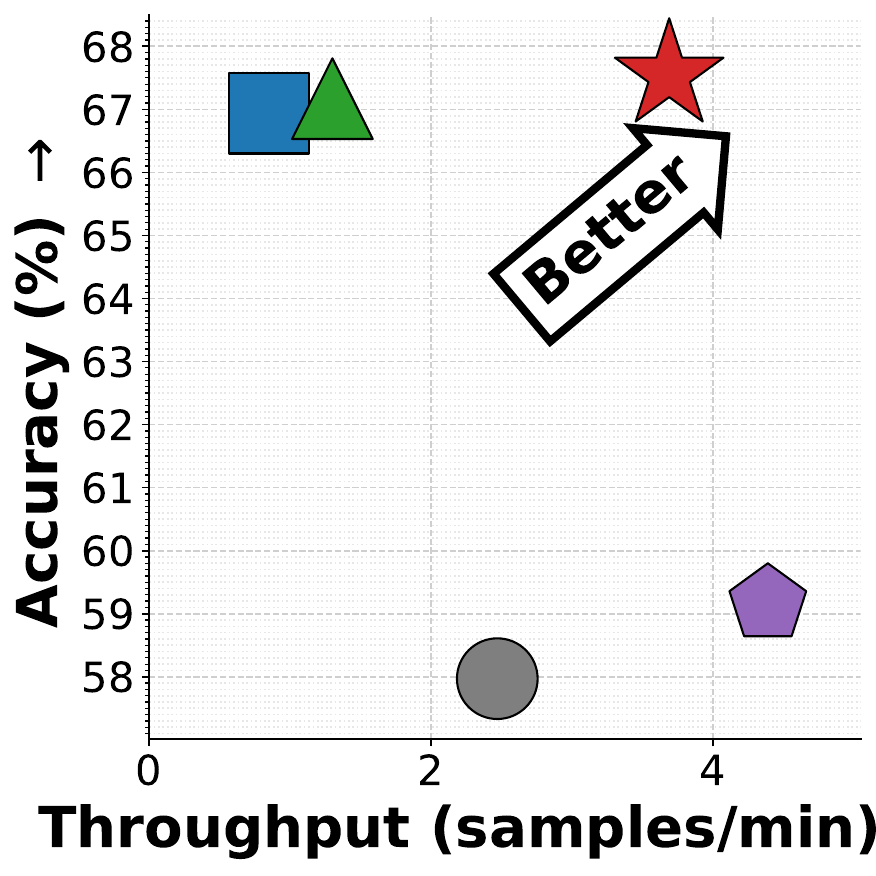}
		\vspace{-0.3cm}
		\caption{Qwen3-VL-8B}
	\end{subfigure}%
	\vspace{-0.1cm}
	\caption{VideoMM sets a new accuracy--efficiency frontier on LongVideoBench, achieving a $6.13\times$ inference speedup with a $7.4\%$ accuracy gain over the vanilla baseline, and a $2.73\times$ speedup over FlexSelect without accuracy loss.
    }
	\vspace{-0.1cm}
	\label{fig:main_performance}
\end{figure}

To mitigate this, existing research has focused on reducing the number of visual tokens via auxiliary models prior to full inference, yet these approaches face a \textit{fundamental dilemma}. 
Lightweight encoder-driven methods \cite{bolya2022tome, liu-etal-2025-video, yang2024visionzip} are efficient but often discard semantically critical information due to limited semantic awareness, while MLLM-driven selection \cite{zhang2025flexselectflexibletokenselection} improves accuracy at the cost of substantial computational overhead. 
As a result, existing model-centric downsizing methods struggle to reconcile the trade-off between accuracy and efficiency in token selection. 
We argue that this bottleneck arises from {\it a more fundamental redundancy}: 
in long-video understanding, different stages of inference require fundamentally different levels of perceptual fidelity. 

While high-fidelity visual details are essential for precise semantic understanding, they are largely unnecessary for the preliminary task of identifying which regions are relevant. Consequently, performing semantic filtering directly on original high-resolution visual tokens is inherently wasteful.

Drawing inspiration from the coarse-to-fine nature of human perception-where one first grasps the global context (macro) before focusing on specific details (micro)—we propose {\bf VideoMM}, a macro-micro paradigm that decouples semantic localization from fine-grained reasoning. 
VideoMM initially identifies relevant regions using a lightweight {\it Macro Proxy} derived from downscaled frames. Subsequently, it employs a consensus mechanism to govern adaptive computation: high-resolution {\it Micro Tokens} are selectively activated only when the necessity for fine-grained inference is confirmed. By strictly adhering to this coarse-to-fine progression, VideoMM ensures that heavy computation is reserved solely for regions requiring granular scrutiny, thereby minimizing redundant computation without compromising semantic fidelity. Extensive evaluations across three benchmarks validate the effectiveness of our approach. Notably, as shown in Figure~\ref{fig:main_performance}, VideoMM achieves an average speedup of 6.13× and an accuracy gain of 7.0\% over vanilla baselines and  accelerates state-of-the-art methods by over 2.73× with  comparable accuracy on the challenging LongVideoBench.

\noindent\textbf{To summarize, our contributions are as follows:} \begin{itemize} 

\item We identify a \textit{fundamental dilemma} in existing token reduction methods rooted in model downsizing, which struggle to balance accuracy with efficiency. {To resolve this, we propose a paradigm shift from model scale to \textit{perceptual granularity}, exploiting the coarse-to-fine nature of visual information to reconcile the conflict between selection precision and overhead.}

\item  We propose VideoMM, a macro-micro paradigm mimicking human coarse-to-fine perception: \textit{Grouped Selection with Macro-view Proxy} decouples token selection from dense processing via lightweight proxies, while \textit{Adaptive Macro-Micro Inference} dynamically recruits specific details only when necessary for ambiguous cases.

\item Extensive evaluations validate the effectiveness of VideoMM, achieving a 6.1× speedup and a 7.0\% accuracy gain on LongVideoBench. Through further analysis of macro-micro variants, we demonstrate the intrinsic efficacy of {this perceptual granularity-oriented paradigm}, establishing a foundational and promising direction that moves beyond the existing model-centric perspective.

\end{itemize}
\section{Related Works}

Given the escalating visual token counts in video MLLMs, token reduction is essential for efficient inference. Current strategies typically employ an additional downscaling model to prune unimportant tokens prior to full inference. Depending on the downscaling model used, these strategies fall into two main categories: \textit{(1) lightweight Encoder-Driven methods \cite{bolya2022tome, yang2024visionzip, liu-etal-2025-video, tao2025dycoke}} selects tokens using only the ViT encoder. 
For example, VisionZip \cite{yang2024visionzip} first selects a subset of dominant visual tokens using ViT attention and then merges the remaining less important tokens through averaging based on embedding similarity, thereby achieving effective token reduction. In contrast, VidCom \cite{liu-etal-2025-video} uses the diversity of ViT embeddings to represent frame-wise importance, subsequently applying different token compression ratios to different frames. However, these methods are prone to erroneous pruning, as lightweight ViTs lack the capacity to fully capture complex video semantics. \textit{(2) Recent heavyweight MLLM-driven methods \cite{chen2024imageworth12tokens, zhang2024sparsevlm, zhang2025flexselectflexibletokenselection}} further leverage a small-scale MLLM  to more accurately detect and preserve essential visual tokens. For instance, the current SOTA method, FlexSelect \cite{zhang2025flexselectflexibletokenselection}, employs a reduced-size MLLM to identify significant tokens via cross-modal attention, subsequently feeding only these selected tokens and the text query into the target MLLM for inference. While this approach improves accuracy, it remains computationally expensive; the overhead introduced by the auxiliary MLLM effectively negates the latency gains achieved through token reduction. 

Overall, the prevailing emphasis on the downscaling model’s size for fast token reduction results in an accuracy–efficiency dilemma. In contrast, VideoMM introduces a paradigm shift by shifting the focus from model scale to perceptual granularity. By constructing a macro-level proxy for fast token reduction, VideoMM substantially reduces the computational overhead of this process—even when using a MLLM to guide compression—without sacrificing semantic depth. Empirically, VideoMM achieves faster inference than lightweight encoder-driven methods while preserving the accuracy advantages of MLLM-driven strategies.~\footnote{Broader related works are discussed in Appendix \ref{apdx:related}}

\begin{figure*}[t!]
  \centering
  \includegraphics[width=\linewidth]{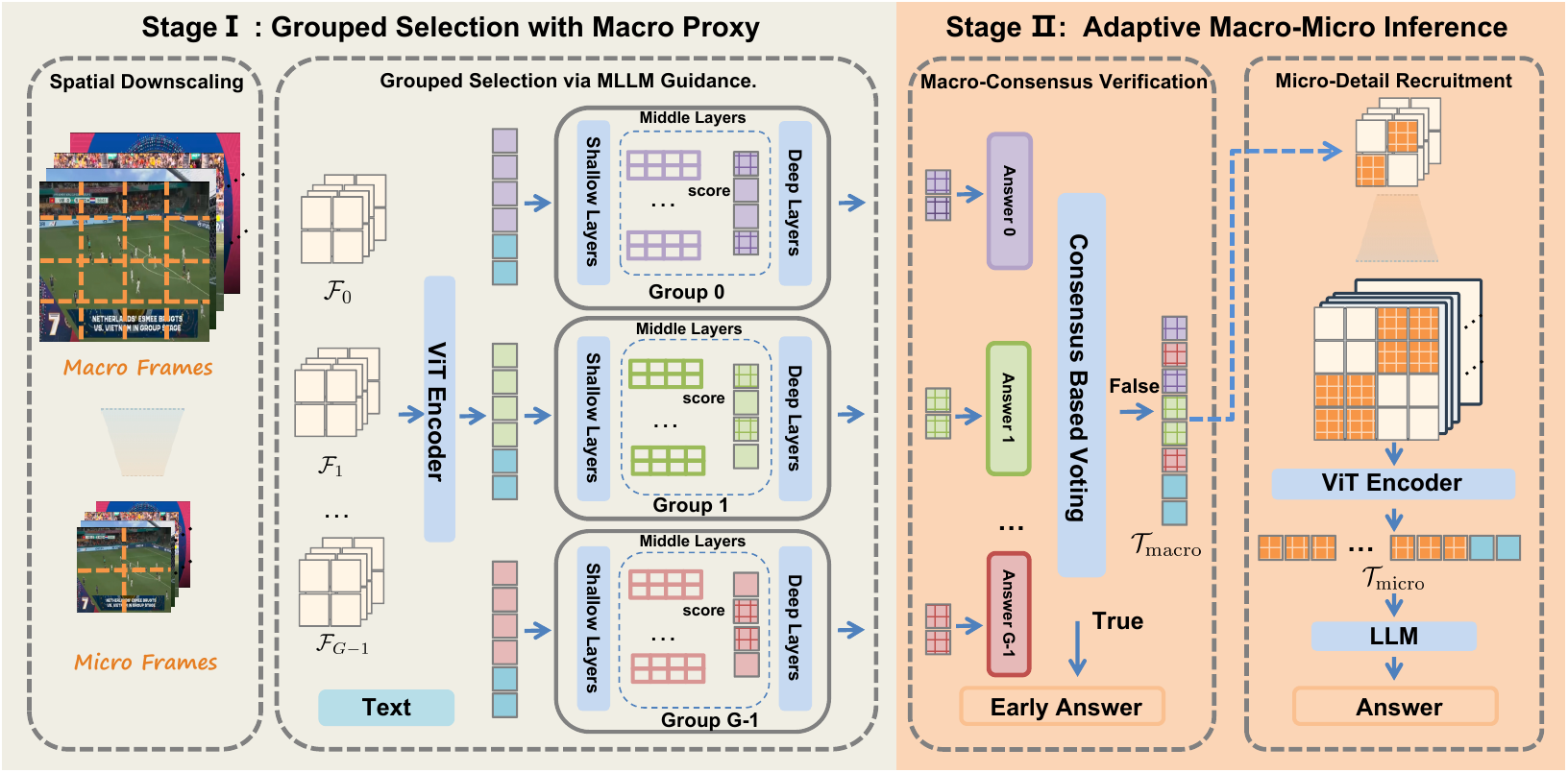}
  \caption{VideoMM Framework. (a) Stage 1: Grouped Selection with Macro-view Proxy efficiently identifies informative regions. (b) Stage 2: Adaptive Macro-Micro Inference assesses semantic sufficiency via consensus, integrating Micro-view video only when necessary.}
  \label{fig:framework}
\end{figure*}

\section{Methods}
We consider standard MLLMs generally consisting of two primary components: a vision transformer encoder (ViT), and a large language model $M$. Given an input video $\mathbf{V} \in \mathbb{R}^{T \times H \times W \times 3}$, a ViT encoder directly processes the frames into a sequence of visual tokens:
\begin{equation}
    \mathbf{X}_V = \text{ViT}(\mathbf{V}) \in \mathbb{R}^{N \times D},
\end{equation}
where $N = T \times H_p \times W_p$ is the total number of visual tokens, $T$ is the number of sampled video frames, $H_p, W_p$ denotes the patched height and patched width respectively, and $D$ is the hidden dimension of the ViT.
Finally, the visual tokens $\mathbf{X}_V$ are concatenated with the text query tokens $\mathbf{X}_Q$. The LLM $M$ generates the response $\mathbf{Y} = \{y_i\}_{i=1}^{L}$ autoregressively by modeling the conditional probability:
$$
p(\mathbf{Y} \mid \mathbf{X}_V, \mathbf{X}_Q) = \prod_{i=1}^{L} p(y_i \mid \mathbf{X}_V, \mathbf{X}_Q, y_{<i}),
$$
where $y_{<i}$ represents the tokens generated prior to step $i$.

\subsection{Overall Architecture of VideoMM}

We present VideoMM, a framework that operationalizes the coarse-to-fine nature of human perception to resolve the efficiency-accuracy dilemma in long-form video understanding. The core insight driving our architecture is that while long videos contain massive tokens, semantic density is sparse: a global macro-view is often sufficient for general reasoning, whereas micro details are only necessary for specific, ambiguous segments. As illustrated in Figure~\ref{fig:framework}, VideoMM implements this philosophy through pipelined collaboration between two distinct stages. (1) \textit{Grouped Selection with Macro Proxy}: This stage acts as a lightweight semantic scout. By operating on a spatially downscaled proxy, it swiftly isolates semantically relevant temporal regions under MLLM guidance, thereby circumventing the prohibitive computational cost of processing raw high-resolution frames. (2) \textit{Adaptive Macro-Micro Inference}: This stage mimics the human cognitive process of attentional zooming, where fine-grained details are recruited only when global perception proves ambiguous. Instead of indiscriminately processing high-resolution tokens, the model assesses the sufficiency of macro-view via a consensus mechanism. If the global context yields a confident answer, inference terminates early; the model activates high-resolution micro tokens only when necessary to resolve uncertainty. This hierarchical design ensures that computational resources are allocated precisely where needed, achieving substantial inference acceleration without compromising the model's ability to resolve subtle visual details.

\subsection{ Stage I: Grouped Selection with Macro Proxy}
To balance the precision of MLLM-guided token selection with computational efficiency, we operate on a spatially downscaled representation of the input video. 
This strategy is grounded in the insight that downscaling significantly reduces data volume while preserving the spatial consistency of critical features. By identifying salient regions on this lightweight proxy, we substantially minimize the computational overhead of the selection process. Please refer to Algorithm~\ref{alg:stage1} for the Stage I pseudocode.

\begin{algorithm}[t]
\caption{Stage I: Grouped Selection with Macro Proxy}
\label{alg:stage1}
\begin{algorithmic}[1]
\Require Video $\mathbf{V} \in \mathbb{R}^{T \times H \times W \times 3}$, Query  $Q$, Scale $k$, Groups $G$, Token budget $B$
\Ensure Macro-level token set $\mathcal{T}'$

\Statex \textbf{1. Spatial Downscaling to Macro Proxy}
\State $\mathbf{V}' \gets \text{Downscale}(\mathbf{V}, k)$ \Comment{$\mathbf{V}' \in \mathbb{R}^{T \times \frac{H}{k} \times \frac{W}{k} \times 3}$}
\State $\mathbf{X}_{\mathbf{V}'} \gets \mathrm{ViT}(\mathbf{V}')$ \Comment{Token count reduced by $k^2$}
\State $\mathcal{T}' \gets \emptyset$

\Statex \textbf{2. Grouped Selection via MLLM Guidance}
\State Interleave frames $f_{1:T}$ into $G$ groups $\{\mathcal{F}_j\}$

\For{each group $\mathcal{F}_j$}
    \State Extract cross-modal $A^{(l,h)}_{q,i}$ for visual token $i$
    \State $s_i = \max_{l \in \mathcal{L}_{\text{mid}}}  \max_{h} \max_{q} A^{(l,h)}_{q,i}$
    \State $s_f \gets \sum_{i \in f} s_i$ for each frame $f \in \mathcal{F}_j$
    \State \textcolor{gray}{// Retain top-50\% frames}
    \State $\mathcal{F}_j' \gets \operatorname{TopRatio}(\mathcal{F}_j, \alpha, \text{key}=s_f)$ 
    \State \textcolor{gray}{// Select top tokens from retained frames}
    \State $\mathcal{T}_j' \gets \operatorname{TopK}(\{\mathbf{X}_{V'} \in \mathcal{F}_j'\}, B/G, \text{key}=s_i)$ 
    
    \State $\mathcal{T}' \gets \mathcal{T}' \cup \mathcal{T}_j'$
\EndFor
\State \Return $\mathcal{T}'$
\end{algorithmic}
\end{algorithm}

\textbf{Spatial Downscaling to Macro Proxy.} 
Given an input video $\mathbf{V} \in \mathbb{R}^{T \times H \times W \times 3}$, we generate a \emph{macro} proxy $\mathbf{V}' \in \mathbb{R}^{T \times \frac{H}{k} \times \frac{W}{k} \times 3}$ by uniformly downscaling spatial dimensions by a factor~\footnote{We set the downscaling factor to 2 in the main experiments, while exploring a more aggressive factor of 3 in Section \ref{sc:ana}.} of $k$. This transformation reduces the token count by $k^2$:
\[
\begin{aligned}
\mathbf{X}_{\mathbf{V}'} &= \mathrm{ViT}(\mathbf{V}') \in \mathbb{R}^{N/k^2 \times D} 
\end{aligned}
\]
By replacing $\mathbf{V}$ with $\mathbf{V}'$ for selection, we drastically reduce the number of visual tokens by a factor of $k^2$ which facilitates subsequent MLLM-guided selection while preserving the global layout of salient features.

\begin{figure*}[t]
  \centering
  \includegraphics[width=1.0\linewidth]{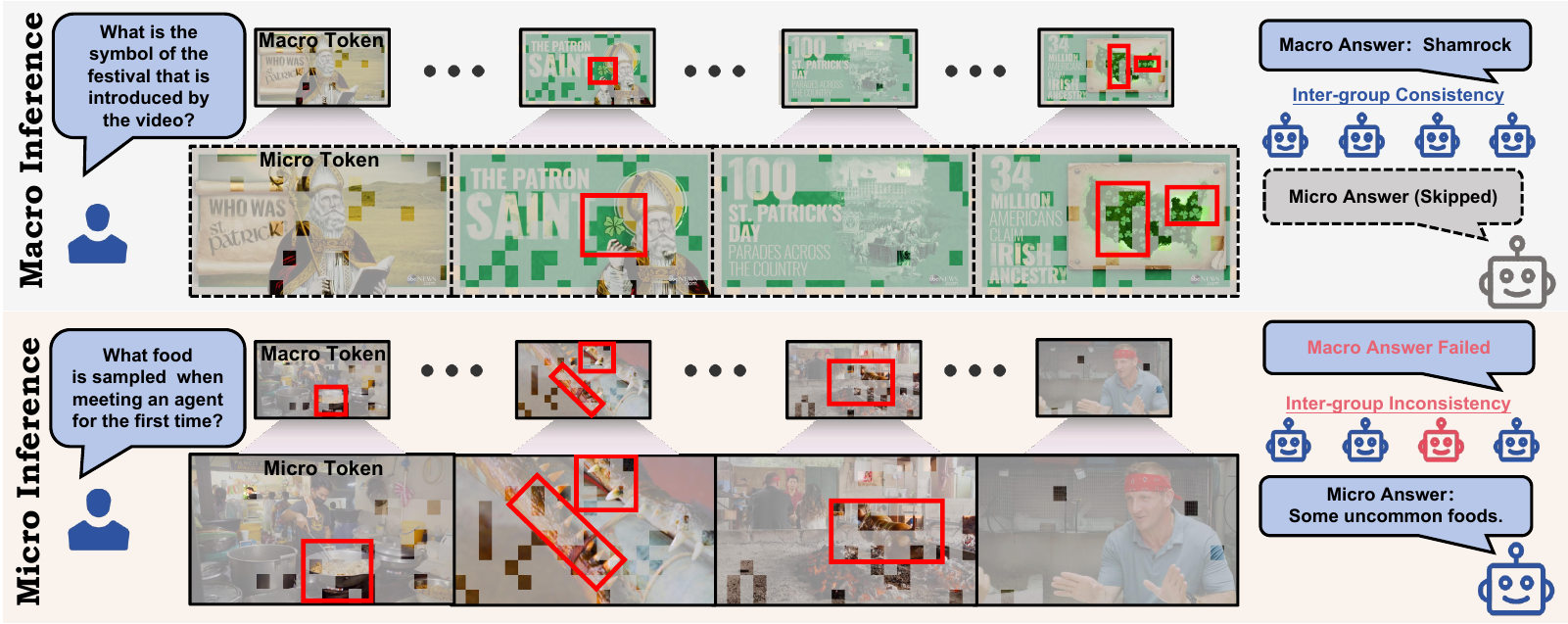}
  \caption{Illustration of Stage II: Adaptive Macro-Micro Inference. \textbf{Top:} In unambiguous cases, the voting mechanism across macro groups reaches a consensus, enabling the model to generate an immediate response. \textbf{Bottom:} When macro-views diverge (indicating ambiguity), VideoMM maps the selected macro regions to micro-level tokens for high-resolution refinement.}
  \label{fig:example}
\end{figure*}

\textbf{Grouped Selection via MLLM Guidance.}
To identify informative tokens within context constraints, we process the macro-view proxy $\mathbf{V}'$ using an interleaved grouping strategy. We partition the $T$ frames (denoted as $f_1, f_2, \dots, f_T$) into $G$ distinct groups:

\begin{equation}
\mathcal{F}_j = \{ f_i \mid i \in [1, T], \ i \equiv j \pmod{G} \},
\end{equation}
where $j \in \{0, \dots, G-1\}$. This decomposition ensures that each group functions as a sparse, global representation, enabling the MLLM to form an independent yet comprehensive understanding of the video content within each group.

Within each group, we quantify the semantic relevance of visual tokens based on the cross-modal attention weights from the text query $\mathbf{X}_Q \in \mathbb{R}^{N_Q \times D}$. Specifically, for a head $h$ in layer $l$, let $\mathbf{W}_Q^{(l,h)}, \mathbf{W}_K^{(l,h)} \in \mathbb{R}^{D \times d_h}$ be the projection matrices, where $d_h$ is the head dimension. The attention weights from text to visual tokens are computed as:
{\small
\begin{equation}
\begin{aligned}
\mathbf{A}^{(l,h)} &\in \mathbb{R}^{N_Q \times N/k^2} \\
&= \operatorname{Softmax}\left(
\frac{(\mathbf{X}_Q \mathbf{W}_Q^{(l,h)})(\mathbf{X}_{V'} \mathbf{W}_K^{(l,h)})^\top}{\sqrt{d_h}}
\right)
\end{aligned}
\end{equation}
}
where $\mathbf{A}^{(l,h)}_{q, i} $ represents the attention weight from the $q$-th text token to the $i$-th visual token. 
We focus on intermediate layers $\mathcal{L}_{\text{mid}}$, as recent studies on long-context LLM inference~\cite{wang2025sparsemm, feng2024ada, xiao2024duoattentionefficientlongcontextllm, wu2024retrievalheadmechanisticallyexplains} suggest that semantic retrieval is predominantly localized in the middle layers.~\footnote{While profiling specific retrieval layers even heads~\cite{zhang2025flexselectflexibletokenselection, fu2024headkv} will yield additional gains, we opted for a fixed selection of three intermediate layers to ensure both simplification and broad applicability.} Inspired by the design of the recent worst-case importance indicator ~\cite{feng2025taming}, we employ maximum aggregation to distill these signals into a robust importance score $s_i$

\begin{equation}
    s_i = \max_{l \in \mathcal{L}_{\text{mid}}}  \max_{h} \max_{q} A^{(l,h)}_{q,i} 
\end{equation}

With the relevance scores $s_i$ established, we implement a hierarchical selection process to filter redundancy at both the frame and token levels. First, at the frame level, we aggregate the importance of each frame $f$ by summing its constituent token scores, denoted as $S_f = \sum_{i \in f} s_i$. For each group $\mathcal{F}_j$, we retain the top $\alpha$ fraction of frames (where $\alpha = 0.5$ in our experiments) with the highest $S_f$ values, forming a refined subset $\mathcal{F}_j'$.\footnote{This $\alpha$-fraction frame selection primarily aims to reduce the computational overhead of re-encoding micro tokens using the ViT encoder in the subsequent stage.} Subsequently, at the token level, we select the $B$ tokens uniformly from each frame groups to construct the final macro-level token set $\mathcal{T}'$:
\begin{equation}
\small
\mathcal{T}' = \bigcup_{j=0}^{G-1} \operatorname{TopK}(\{\mathbf{X}_{V'} \in \mathcal{F}_j'\}, B/G, \text{key}=s_i)
\end{equation}

\begin{algorithm}[t]
\caption{Stage II: Adaptive Macro-Micro Inference}
\label{alg:stage2}
\begin{algorithmic}
\Require Macro token set $\mathcal{T}'$ (from Stage I), Query $Q$, Downscale factor $k$,  Frame Group $\mathcal{F}_j$, Model $\mathcal{M}$
\Ensure Final Answer $R$

\State \textbf{1. Macro-Consensus Verification}
\State $Y_{\text{response}} \gets  \{\mathcal{M}(\mathcal{T}', Q)\}$

\For{each group $\mathcal{F}_j$}
  \State Extract a subset $\mathcal{T}'_j \subseteq \mathcal{T}'$ belonging to group $\mathcal{F}_j$
  \State $ans_j \gets \mathcal{M}(\mathcal{T}_j, Q)$ \Comment{Generate answer per group}
  \State Add $ans_j$ to $Y_{\text{response}}$
\EndFor

\If{All responses in $Y_{\text{response}}$ are consistent}
  \State \Return the consensus result from $Y_{\text{response}}$ 
\EndIf

\State \textbf{2. Micro-Detail Recruitment}
    \State Obtain raw frames $\mathbf{\hat{V}}$ associated with $\mathcal{T}'$
    \State Encode micro features: $\mathbf{X}_{\hat{V}} \gets \text{ViT}(\mathbf{\hat{V}})$
     \State \textcolor{gray}{// Macro-to-Micro token mapping via Eq. \ref{equ: mapping}}
    \State $\mathcal{T} \gets \text{Gather}(\mathbf{X}_{\hat{V}}, \mathcal{T}')$ 
    \State \Return $\mathcal{M}(\mathcal{T}, Q)$
\end{algorithmic}
\end{algorithm}

\subsection{Stage II: Adaptive Macro-Micro Inference}

While Stage I identifies \textit{where} to look, Stage II determines \textit{how deep} to look. As illustrated in Figure~\ref{fig:example}, this stage is grounded in the principle of computational efficiency: most video queries can be answered via global semantic cues (Macro), while only a fraction require high-resolution scrutiny (Micro). To exploit this, VideoMM employs an adaptive granularity inference path, prioritizing low-cost macro inference and escalating to high-cost micro reasoning only when ambiguity is detected. Please refer to Algorithm~\ref{alg:stage2} for the Stage II pseudocode.

\textbf{Macro-Consensus Verification.}
We employ a \textit{Consensus-Based Voting} mechanism \cite{wang2023selfconsistencyimproveschainthought} to evaluate the sufficiency of the macro-view. Each group $\mathcal{F}_j$, alongside the jointly selected macro token set $\mathcal{T}'$ (treated as an additional independent group), processed independently, acts as a weak estimator generating a preliminary response. We posit that prediction invariance across disjoint temporal subsets serves as a robust proxy for confidence. If all independent groups converge on a consensus, macro-level semantics are deemed sufficient \footnote{Other voting strategies are also applicable; notably, majority voting boosts throughput with little accuracy drop (Section~\ref{sc:ana}).}. In such cases, the model terminates inference early using the consensus result, bypassing the overhead of micro-level processing. \footnote{Given that prefilling is completed in Stage I and each group contains only a few macro Tokens, the overhead of this consensus voting is negligible relative to its substantial benefits.}

\textbf{Micro-Detail Recruitment.}
When macro-views diverge—indicating ambiguity and the need for finer detail—, VideoMM activates micro-level refinement. 
This process necessitates encoding the relevant raw video frames into micro tokens:
{\small
\begin{equation}
\mathbf{X}_{\hat{V}} = \text{ViT}(\mathbf{\hat{V}}) \in \mathbb{R}^{\alpha N  \times D}    
\end{equation}
}

where $\mathbf{\hat{V}}$ denotes the raw video input constructed from the frames selected in the first stage, consisting of only the $\alpha$-fraction of the original frames. This effectively controls the computational overhead of ViT re-encoding for these micro tokens. 
We then map the critical macro tokens $\mathcal{T}'$ selected in Stage I back to their corresponding micro-level representations $\mathcal{T}$. Specifically, for any macro token in $\mathcal{T}'$ identified by $(t,h',w')$ in frame and spatial coordinates, it corresponds to a micro-token set from the newly encoded  $\mathbf{X}_{\hat{V}}$ within the same spatial region:
{\small
\begin{equation}
\mathcal{T}_{(t,h,w)} = \left\{ \mathbf{X}_{\hat{V}}[t, h, w] \mid
\begin{cases}
h'  k \le h < (h'+1)  k,\\
w'  k \le w < (w'+1)  k
\end{cases} \right\}.
\label{equ: mapping}
\end{equation}
}

The complete micro token set $\mathcal{T}$ is then the union of these subsets: $\mathcal{T} = \bigcup \mathcal{T}_{(t,h,w)}$, which is then fed into the models for the definitive response. This paradigm balances efficiency and accuracy by using the macro stage as a filter for noise and the micro stage as a specialized solver for detail.

\begin{table*}[t!]
	\centering
	\caption{Performance comparison on video understanding benchmarks. Throughput denotes inference speed in samples per minute. The best two results are \textbf{bolded}. VideoMM demonstrates leading performance in both accuracy and throughput, whereas other methods typically prioritize one over the other.}
	\label{tab:videobench}
	\resizebox{0.99\textwidth}{!}{%
		\setlength{\tabcolsep}{6pt}
        \renewcommand{\arraystretch}{0.9}
		\begin{tabular}{@{}lllllllll@{}}
        
			\toprule
			
			\multirow{2}{*}{Method} & \multicolumn{2}{c}{LongVideoBench} & \multicolumn{2}{c}{VideoMME} & \multicolumn{2}{c}{LVBench} & \multicolumn{2}{c}{Average} \\ 
			\cmidrule(lr){2-3} \cmidrule(lr){4-5} \cmidrule(lr){6-7} \cmidrule(lr){8-9}
			
             & Accuracy  & Throughput  & Accuracy & Throughput  & Accuracy  & Throughput  & Accuracy & Throughput  \\
			
            \midrule
            \multicolumn{9}{c}{Qwen2.5-VL-7B}\\
            \midrule
             Vanilla & 57.67 & 0.77 & 61.56 & 0.85 & 39.57 & 0.65 & 52.93 & 0.76 \\
            \arrayrulecolor{lightgray}
            \midrule
            \arrayrulecolor{black}
            
             VisionZip & 58.23 (↑1.0\%) & 3.63 (4.7x) & 62.14 (↑0.9\%) & 3.84 (4.5x) & 41.38 (↑4.6\%) & 3.10 (4.8x) & 53.92 (↑1.9\%) & 3.52 (4.6x) \\
             VidCom & 54.60 (↓5.3\%) & \textbf{4.27 (5.5x)} & 61.56 (-0.0\%) & \textbf{4.42 (5.2x)} & 40.80 (↑3.1\%) & \textbf{4.32 (6.6x)} & 52.32 (↓1.2\%) & \textbf{4.34 (5.7x)} \\
            \arrayrulecolor{black}
            
             Flexselect & \textbf{64.10 (↑11.1\%)} & 1.83 (2.4x) & \textbf{68.85 (↑11.8\%)} & 1.84 (2.2x) & \textbf{51.45 (↑30.0\%)} & 1.50 (2.3x) & \textbf{61.47 (↑16.1\%)} & 1.72 (2.3x) \\
             \rowcolor{gray!15}VideoMM & \textbf{64.70 (↑12.2\%)} & \textbf{5.16 (6.7x)} & \textbf{68.41 (↑11.1\%)} & \textbf{5.47 (6.4x)} & \textbf{50.94 (↑28.7\%)} & \textbf{4.20 (6.5x)} & \textbf{61.35 (↑15.9\%)} & \textbf{4.94 (6.5x)} \\

            \midrule
        \multicolumn{9}{c}{GLM-4.1V-9B}\\
        \midrule
         Vanilla & 58.98 & 0.51 & 66.67 & 0.58 & 47.90 & 0.42 & 57.85 & 0.50 \\
        \arrayrulecolor{lightgray}
        \midrule
        \arrayrulecolor{black}
        
         VisionZip & 55.35 (↓6.2\%) & \textbf{4.10 (8.0x)} & 59.55 (↓10.7\%) & \textbf{4.30 (7.4x)} & 36.41 (↓24.0\%) & \textbf{3.60 (8.6x)} & 50.44 (↓12.8\%) & \textbf{4.00 (8.0x)} \\
         VidCom & 51.68 (↓12.4\%) & 3.50 (6.9x) & 56.48 (↓15.3\%) & 3.70 (6.4x) & 37.70 (↓21.3\%) & \textbf{2.96 (7.0x)} & 48.62 (↓16.0\%) & 3.39 (6.8x) \\
        \arrayrulecolor{black}
        
         Flexselect & \textbf{60.66 (↑2.8\%)} & 1.47 (2.9x) & \textbf{65.15 (↓2.3\%)} & 1.57 (2.7x) & \textbf{50.16 (↑4.7\%)} & 1.28 (3.0x) & \textbf{58.66 (↑1.4\%)} & 1.44 (2.9x) \\
         \rowcolor{gray!15}VideoMM & \textbf{64.32 (↑9.1\%)} & \textbf{3.77 (7.4x)} & \textbf{67.37 (↑1.0\%)} & \textbf{4.21 (7.3x)} & \textbf{49.52 (↑3.4\%)} & 2.87 (6.8x) & \textbf{60.40 (↑4.4\%)} & \textbf{3.62 (7.2x)} \\

			  \midrule
            \multicolumn{9}{c}{Qwen3-VL-8B}\\
            \midrule
             Vanilla & 66.94 & 0.85 & 71.78 & 0.97 & 54.1 & 0.70 & 64.27 & 0.84 \\
            \arrayrulecolor{lightgray}
            \midrule
            \arrayrulecolor{black}
            
             VisionZip & 57.97 (↓13.4\%) & 2.47 (2.9x) & 65.55 (↓8.7\%) & 2.43 (2.5x) & 40.22 (↓25.7\%) & 1.92 (2.7x) & 54.58 (↓15.1\%) & 2.27 (2.7x) \\
             VidCom & 59.16 (↓11.6\%) & \textbf{4.39 (5.2x)} & 64.59 (↓10.0\%) & \textbf{3.04 (3.1x)} & 44.03 (↓18.6\%) & \textbf{2.39 (3.4x)} & 55.93 (↓13.0\%) & \textbf{3.27 (3.9x)} \\
            \arrayrulecolor{black}
            
             Flexselect & \textbf{67.17 (↑0.3\%)} & 1.30 (1.5x) & \textbf{72.19 (↑0.6\%)} & 1.32 (1.4x) & \textbf{55.20 (↑2.0\%)} & 0.97 (1.4x) & \textbf{64.85 (↑0.9\%)} & 1.20 (1.4x) \\
             \rowcolor{gray!15}VideoMM & \textbf{67.54 (↑0.9\%)} & \textbf{3.69 (4.3x)} & \textbf{71.67 (↓0.2\%)} & \textbf{3.59 (3.7x)} & \textbf{54.36 (↑0.5\%)} & \textbf{2.50 (3.6x)} & \textbf{64.52 (↑0.4\%)} & \textbf{3.26 (3.9x)} \\

			\bottomrule
		\end{tabular}%
	}
\end{table*}

\subsection{ Efficiency Analysis}

To quantify the efficiency gains of VideoMM, we analyze its asymptotic complexity relative to standard MLLM inference. We focus on the self-attention mechanism, which constitutes the primary computational bottleneck in long-video understanding. Let $N$ denote the number of visual tokens in raw high-resolution video. Given that $N$ scales significantly with video duration, we analyze complexity with respect to $N$, treating text sequence length as negligible.

\noindent \textbf{Complexity Derivation.}
Standard global attention generally scales quadratically, i.e., $\mathcal{O}(N^2)$. VideoMM optimizes this via three key parameters: the macro downscaling factor $k$, the number of temporal groups $G$, and the macro-selection budget $B$ (introduced in Stage I) combined with the macro-consensus probability $\alpha$ (governed by Stage II). The expected computational cost of VideoMM, $\mathbb{E}[\Omega_{\text{Ours}}]$, is the sum of two components:
\begin{enumerate}
    \item \textbf{Macro Processing (Stage I \& Stage II-Consensus):} 
    Both \textit{Grouped Selection} (Stage I) and \textit{Macro-Consensus Verification} (Stage II) operate on shared \textit{Macro Proxy}. Here, token count is reduced to $N/k^2$ and divided into $G$ groups. Since attention is restricted within these groups for relevance estimation ($s_i$) and consensus voting, the computational overhead is:
    $ G \cdot \left(\frac{N/k^2}{G}\right)^2 = \frac{N^2}{G k^4} $.
    
    \item \textbf{Micro Processing (Stage II-Refinement):} 
    This term corresponds to the conditional activation of the \textbf{Micro-Detail Recruitment} phase in Stage II. Let $\rho = (B \cdot k^2)/N$ denote the ratio of selected high-resolution tokens. 
    Crucially, this computation is only incurred when the consensus mechanism \textit{fails}. Let $\beta$ denote the probability of successfully reaching a consensus (i.e., early exit). The expected cost is:$ (1-\beta)(\rho N)^2 $.

\end{enumerate}

Summing these terms, the total expected complexity is:
{
\small
\begin{equation}
    \mathbb{E}[\Omega_{\text{Ours}}] = \mathcal{O}\left( N^2 \left[ \underbrace{\frac{1}{G k^4}}_{\text{Macro Proxy}} + \underbrace{(1 - \beta)\rho^2}_{\text{Micro Refinement}} \right] \right).
    \label{equ:advantage}
\end{equation}
}

Equation \ref{equ:advantage} demonstrates the transformation of quadratic dependency into an efficient form via two benefits:
\textit{(1) Grouping and Downscaling:} The term $\frac{1}{G k^4}$ represents the structural cost reduction, where $k$ and $G$ provide quartic and linear complexity drops, respectively. \textit{(2) Adaptive Inference:} The term $(1 - \beta)\rho^2$ captures the gain from adaptive efficiency. 
Empirically, high early-exit rates ($\beta \approx 70\%$) and extreme sparsity ($\rho < 3\%$) confine heavy computation strictly to ambiguous instances, preserving detail with minimal overhead.

\section{Experiments}
\subsection{ Settings}
\textbf{Models.}  To demonstrate the universality of our framework, we conduct evaluations on three leading open-source MLLMs: Qwen2.5-VL-7B \cite{Qwen2.5-VL} and GLM-4.1-VL-9B-Thinking \cite{glm4v} support 64K context windows, while Qwen3-VL-8B \cite{qwen3technicalreport} extends to 256K natively. For GLM-4.1-VL-9B-Thinking, we disable its thinking mode to force direct responses and keep evaluation costs manageable. These models were selected as state-of-the-art for long-context video understanding; furthermore, their dynamic-resolution ViTs naturally support varied input resolutions, inherently aligning with our hierarchical macro-micro paradigm.

\noindent \textbf{Baselines.} 
We benchmark our approach against three representative token reduction methods, categorized by their computational characteristics: \textit{(i) lightweight encoder-driven methods} and \textit{(ii) heavyweight MLLM-driven methods}. For the lightweight category, we evaluate VisionZip \cite{yang2024visionzip}, which utilizes ViT attention for selection, and VidCom~\cite{liu-etal-2025-video}, which employ token uniqueness for compression. For the heavyweight category, we utilize FlexSelect \cite{zhang2025flexselectflexibletokenselection} as a representative baseline. 
To ensure a fair comparison and strictly control computational costs, we adopt a unified token budget across all baseline categories. Specifically, for both lightweight encoder-driven methods~\cite{yang2024visionzip,liu-etal-2025-video} and the heavyweight MLLM-driven FlexSelect~\citep{zhang2025flexselectflexibletokenselection}, we retain a fixed 8,192 visual tokens, following common practice. For VideoMM ($G=4, k=2$), we align our Micro token budget with this unified 8,192 limit; consequently, the Macro budget is set to 2,048 ($8,192/4$).

\noindent \textbf{Benchmarks.} 
Our evaluation employs two widely adopted long-video benchmarks: LongVideoBench\cite{wu2024longvideobench}, focusing on fine-grained retrieval and complex reasoning within long-context videos, and LVBench\cite{wang2024lvbench} for testing the capability over hour-long video analysis. Furthermore, we assess multi-scale temporal understanding using VideoMME\cite{fu2025video}. We sample videos uniformly at 1 FPS, up to a maximum of 512 frames.  All evaluations use the LMMs-Eval framework\cite{zhang2024lmmsevalrealitycheckevaluation} following standard protocols.

\begin{table}[t]
	\centering
	\caption{Comparison on long-video benchmarks under comparable throughput (FlexSelect reduced to 256 frames; Acc.: Accuracy (\%); Thpt.: Throughput (samples/min)) }
	\label{tab:comparison_fs_long}
    \footnotesize
	\setlength{\tabcolsep}{3pt}
	\renewcommand{\arraystretch}{0.8}
	\begin{tabular}{@{}lcccccc@{}}
		\toprule
		\multirow{2}{*}{Method} 
        & \multicolumn{2}{c}{LongVideoBench} 
        & \multicolumn{2}{c}{LVBench} 
        & \multicolumn{2}{c}{Average} \\
		\cmidrule(lr){2-3} \cmidrule(lr){4-5} \cmidrule(lr){6-7}
		& Acc. & Thpt. & Acc. & Thpt. & Acc. & Thpt. \\
        \midrule

		\multicolumn{7}{c}{\textit{Qwen2.5-VL-7B}}\\
		\midrule
        Vanilla (512) & 57.67 & 0.77 & 39.57 & 0.65 & 48.62 & 0.71 \\
        \arrayrulecolor{lightgray}\midrule
		\arrayrulecolor{black}
        FlexSelect (256) & 62.30 & 3.27 & 46.93 & 2.87 & 54.62 & 3.07 \\
        VideoMM (512) & \textbf{64.70} & \textbf{5.16} & \textbf{50.94} & \textbf{4.20} & \textbf{57.82} & \textbf{4.68} \\

        \midrule
		\multicolumn{7}{c}{\textit{GLM-4.1V-9B}}\\
		\midrule
        Vanilla (512) & 58.98 & 0.51 & 47.90 & 0.42 & 53.44 & 0.47 \\
        \arrayrulecolor{lightgray}\midrule
		\arrayrulecolor{black}
        FlexSelect (256) & 63.13 & 3.35 & 48.16 & \textbf{2.97} & 55.65 & 3.16 \\
        VideoMM (512) & \textbf{64.32} & \textbf{3.77} & \textbf{49.52} & 2.87 & \textbf{56.92} & \textbf{3.32} \\

		\midrule
		\multicolumn{7}{c}{\textit{Qwen3-VL-8B}}\\
		\midrule
        Vanilla (512) & 66.94 & 0.85 & 54.10 & 0.70 & 60.52 & 0.78 \\
        \arrayrulecolor{lightgray}\midrule
		\arrayrulecolor{black}
        FlexSelect (256) & 65.82 & \textbf{3.70} & 50.36 & \textbf{3.06} & 58.09 & \textbf{3.38} \\
        VideoMM (512) & \textbf{67.54} & 3.69 & \textbf{54.36} & 2.50 & \textbf{60.95} & 3.10 \\

		\bottomrule
	\end{tabular}

\end{table}

\begin{figure}[t]
    \centering
    \includegraphics[width=0.9\linewidth]{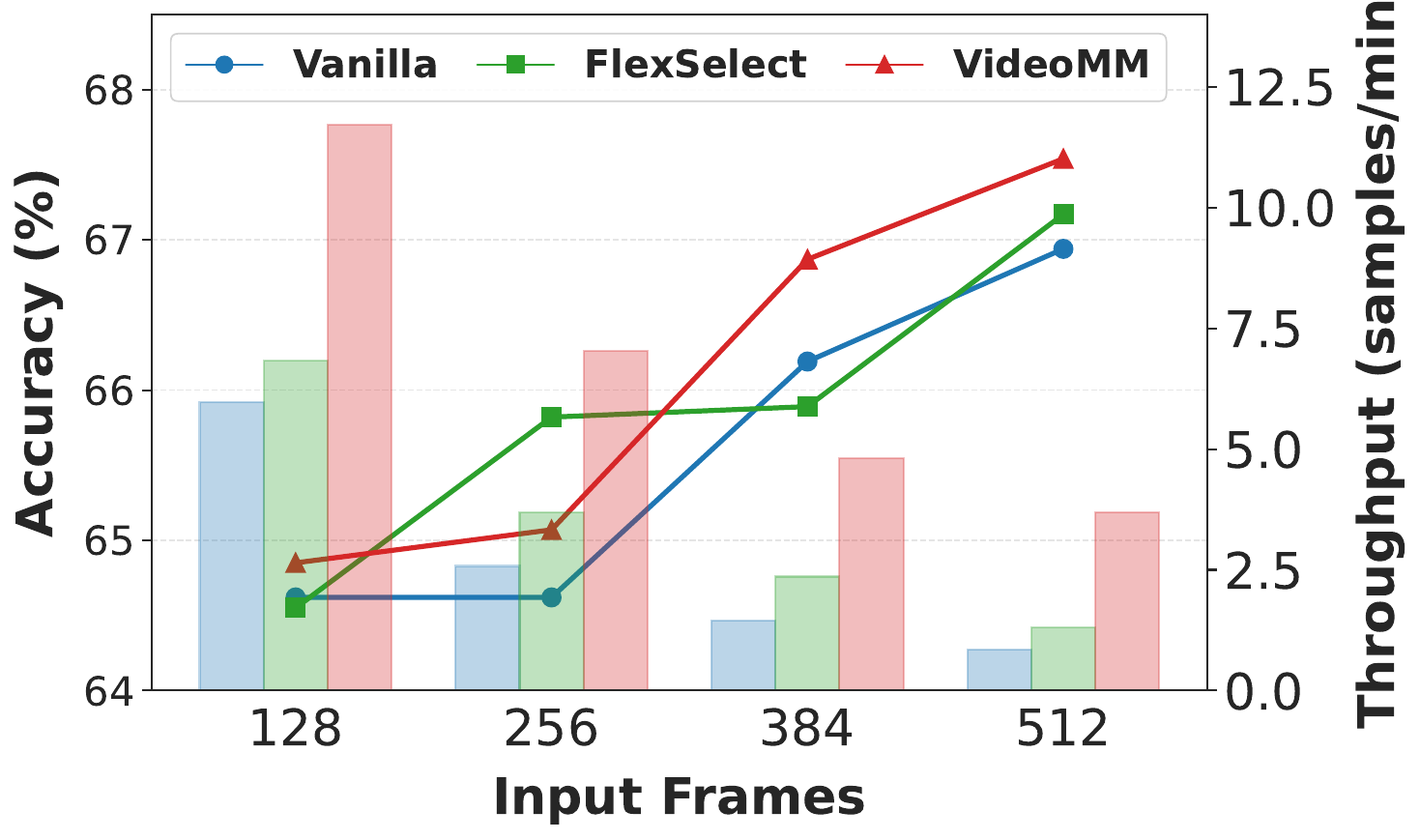}
    
    \caption{Impact of input frame counts on LongVideoBench using Qwen3-VL-8B. The \textbf{solid lines} represent accuracy (left axis), while the \textbf{transparent bars} represent throughput (right axis).}
    \vspace{-0.2cm} 
    \label{fig:input_frame} 
    \vspace{-0.2cm} 
\end{figure}

\subsection{Main Results}

Table~\ref{tab:videobench} presents a comparative evaluation of accuracy and throughput for VideoMM against various token compression methods across multiple models and benchmarks. Overall, \textit{Encoder-driven methods} excel in throughput, whereas \textit{MLLM-driven methods} achieve higher accuracy. In contrast, VideoMM demonstrates superiority in both accuracy and efficiency, enabled by its novel macro-micro perceptual granularity. Detailed performance on the three benchmarks are provided in the appendix.

\noindent \textbf{Comparison with {Encoder-Driven Methods}.}

By leveraging macro-level MLLM-guided token selection, VideoMM significantly outperforms lightweight encoder-driven methods (e.g., VisionZip, VidCom) in both accuracy and throughput.
For instance, on Qwen3-VL-8B and Qwen2.5-VL-7B, VideoMM attains average scores of 64.52 and 61.35 across three benchmarks, substantially outperforming VidCom’s 55.93 and 52.32, respectively.Moreover, even in terms of throughput—where encoder-driven methods typically excel—VideoMM achieves superior performance due to its macro-micro paradigm. On the same models, VideoMM reaches average throughputs of 3.26 and 4.94 samples/min, delivering comparable or superior overall efficiency compared to VidCom’s 3.27 and 4.34.

\noindent \textbf{Comparison with {MLLM-Driven methods.}} 
By employing Adaptive Macro-Micro Inference, VideoMM achieves a substantial boost in throughput while preserving accuracy on par with FlexSelect. For example, on Qwen3 and GLM, VideoMM achieves average scores of 64.52 and 60.40 across three benchmarks, close to FlexSelect’s 64.85 and 58.66, respectively. Crucially, VideoMM demonstrates a significant throughput advantage: on Qwen3, it reaches 3.26 samples/min—surpassing FlexSelect’s 1.20—which corresponds to a \textit{3.9× speedup} over the vanilla. This gap is even more pronounced on GLM-4.1V-9B, where VideoMM achieves 3.62 samples/min compared to FlexSelect’s 1.44, yielding a \textit{7.2× speedup} over vanilla.  In real-world long-video understanding, balancing accuracy and throughput often necessitates adjusting input frame counts to meet deployment constraints. 
Thus, in Table \ref{tab:comparison_fs_long}, we focus specifically on two long-video benchmarks (LongVideoBench and LVBench), employing frame reduction on FlexSelect (to 256 frames) to compare accuracy under iso-throughput conditions.  
Overall, our method consistently achieves higher accuracy at comparable throughput. For instance, with Qwen2.5-VL-7B, VideoMM outperforms FlexSelect by 4.7 point in average accuracy (59.32 vs. 54.62) while simultaneously increasing throughput. This advantage extends to GLM-4.1V-9B and Qwen3-VL-8B, where our method maintains higher accuracy at comparable inference speeds.

\begin{figure}[t]
	\centering
    \includegraphics[width=0.8\linewidth]{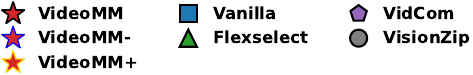}
	\begin{subfigure}{\linewidth}
		\centering
		\includegraphics[width=\linewidth]{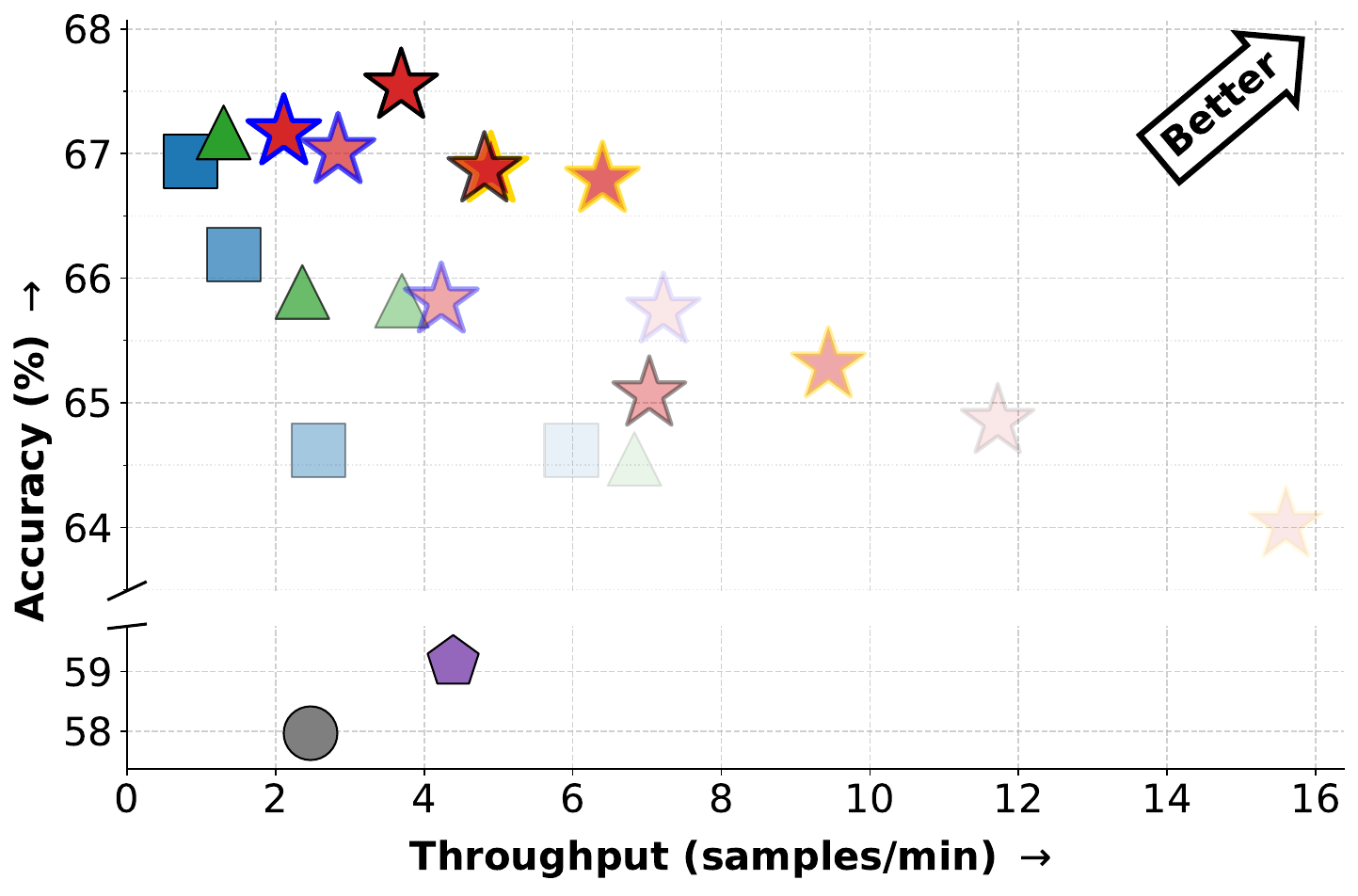}
	\end{subfigure}%
	\vspace{-0.1cm}
	\caption{Accuracy-efficiency trade-offs across VideoMM family variants (Qwen3-VL-8B, LongVideoBench). Legend colors from dark to light correspond to frame counts of 512, 386, 256, and 128, respectively. The VideoMM family establishes a superior Pareto frontier, validating the intrinsic efficacy of the proposed macro-micro  paradigm.}
	\vspace{-0.3cm}
	\label{fig:ablation}
\end{figure}

\begin{table}[t]
	\centering
	\caption{Early-exit statistics $\beta$ (LongVideoBench, Qwen3-VL-8B).}
	\vspace{-0.1cm}
	\label{tab:beta}
	\small
	\setlength{\tabcolsep}{3pt}
	\renewcommand{\arraystretch}{1.0}
	\begin{tabular}{@{}lllll@{}}
		\toprule
		Frame count & 128 & 256 & 384 &  512  \\
        \midrule
        \arrayrulecolor{black}
		VideoMM$^-$       & $0\%$ & $0\%$ & $0\%$ & $0\%$ \\
		VideoMM           & $65.7\%$ & $69.6\%$  & $70.8\%$ & $73.3\%$ \\ 
        VideoMM$^+$       & $88.6\%$ & $89.3\%$ & $88.6\%$ & $88.6\%$  \\
		\bottomrule
	\end{tabular}
	\vspace{-0.3cm}
\end{table}

\subsection{Analysis Experiments} 
\label{sc:ana}

\noindent \textbf{Detail discussion of Adaptive Computation.} VideoMM employs an adaptive computation mechanism within a macro-micro hierarchical framework, dynamically allocating computational resources based on sample complexity.  To investigate the accuracy-efficiency trade-off from this mechanism, we instantiate two variants by modulating the adaptive strategy \footnote{In Appendix \ref{apdx:var1} and \ref{apdx:var2}, we explore additional variants within the Macro-Micro paradigm to further enhance efficiency.}: \textit{(1) VideoMM$^+$ (Efficiency-Prioritized):} This variant adopts a majority-voting strategy for Macro-Consensus Verification, allowing early exit upon majority agreement. This significantly increases the probability of early termination at the cost-efficient macro level compared to the standard VideoMM, which requires unanimous consensus. \textit{(2) VideoMM$^-$  (Non-Adaptive Ablation):} Conversely, this variant enforces all samples to undergo fine-grained inference at the micro level, completely disabling the adaptive computation mechanism. Figure \ref{fig:ablation} illustrates the accuracy-efficiency trade-off across varying frame inputs, while Table \ref{tab:beta} details the early-exit statistics $\beta$. Overall, as we transition from VideoMM$^-$ to VideoMM and finally VideoMM$^+$, both the early-exit probability $\beta$ and inference speed progressively increase, though potentially at the cost of marginal accuracy drops.  Considering both accuracy and efficiency,  both adaptive variants (VideoMM and VideoMM$^+$) yield a superior trade-off compared to the non-adaptive VideoMM$^-$, occupying the optimal upper-right region of the plot. Between the two, the choice depends on specific application constraints. 
{
For instance, with 512-frame inputs, VideoMM prioritizes higher accuracy (67.54) with moderate throughput (3.69), whereas VideoMM$^+$ achieves significantly higher throughput (4.90) with slightly lower accuracy (66.87). 
} Overall, the VideoMM family establishes a new optimal frontier on accuracy-throughput landscape. This success validates the intrinsic efficacy of macro-micro perceptual granularity paradigm, offering a novel direction that moves beyond the prevailing model-centric perspective.

\noindent \textbf{Influence of Input Frames.} As shown in Figure~\ref{fig:input_frame}, we further evaluate the performance of VideoMM on the challenging LongVideoBench benchmark using varying input frame count to simulate diverse real-world deployment constraints. Typically, increasing the number of frames improves long-video understanding accuracy but incurs a substantial token overhead, thereby significantly reducing throughput. However, VideoMM effectively mitigates the trade-off between efficiency and accuracy, maintaining superior accuracy while markedly improving throughput across all input frame settings. { For instance, with 512 frames, VideoMM attains a superior accuracy of 67.54\% and achieves speedups of 4.3× over vanilla and 2.8× over FlexSelect. Notably, the throughput of VideoMM at 512 frames is comparable to that of FlexSelect at only 256 frames.} Consequently, in practical applications, Video-MM enables the efficient processing of longer video inputs, thereby unlocking superior understanding capabilities from these extended contexts.

\noindent \textbf{Influence of Downscale factor $k$.} In our main experiments, we set the downscale factor to $k=2$, reducing the first-stage macro proxy tokens to one-quarter of the original count to maintain high accuracy.  Here, we investigate a more aggressive downscaling strategy by setting $k=3$—which reduces tokens to  1/9—to evaluate the impact on the accuracy-throughput trade-off. 
As shown in Table~\ref{tab:factor_k}, increasing downscale factor $k=3$  naturally incurs a slight accuracy cost compared to $k=2$ (65.89 vs. 67.54). However, this yields a substantial gain in computational speed, nearly doubling the throughput from 3.69 to 6.74 samples per second. 
Crucially, this aggressive $k=3$ configuration establishes a superior Pareto frontier for highly efficient long-video understanding. When benchmarked against lightweight Encoder-driven methods (e.g., VisionZip and VidCom), VideoMM with $k=3$ demonstrates comprehensive dominance: it not only retains higher accuracy (65.89 vs. $\approx$64.0) but also achieves drastically higher throughput (7.9×
 vs. $\approx$2.2×). The extreme downscaling strategy remains highly effective primarily because our Stage II consensus mechanism acts as a reliable safety net; when severe perceptual ambiguity is induced by the $k=3$ compression, the resulting macro-level divergence typically triggers a fallback to micro-level refinement, mitigating the risk of catastrophic reasoning failures. This result confirms the robustness of our macro-proxy mechanism even under aggressive downscaling. Consequently, the downscale factor serves as a practical control knob, allowing users to seamlessly transition between high-precision and high-speed regimes to meet diverse deployment constraints.

\begin{table}[t]
	\centering
	\caption{Impact of downscale factor $k$. Evaluated on LongVideoBench using Qwen3-VL-8B. Throughput is measured in samples per min. Best results are highlighted in \textbf{bold}.}
	\label{tab:factor_k}
	\small
    \vspace{-0.1cm}
	\setlength{\tabcolsep}{5pt}
	\renewcommand{\arraystretch}{1.0}
	\begin{tabular}{lll}
		\toprule
		Method & Accuracy & Throughput \\
		\midrule
		Vanilla          & 66.94 & 0.85 \\
        \arrayrulecolor{lightgray} \midrule
        VisionZip        & 57.97 (↓13.4\%) & 2.47 (2.9x) \\ 
        VidCom           & 59.16 (↓11.6\%) & 4.39 (5.2x) \\ 
        Flexselect       & 67.17 (↑0.3\%) & 1.30 (1.5x) \\ 
		\cellcolor{gray!15}VideoMM, k=2     & \cellcolor{gray!15}\textbf{67.54 (↑0.9\%)} & \cellcolor{gray!15}3.69 (4.3x) \\
		\cellcolor{gray!15}VideoMM, k=3    & \cellcolor{gray!15}65.89 (↓1.6\%) & \cellcolor{gray!15}\textbf{6.74 (7.9x)} \\
        \arrayrulecolor{black}
		\bottomrule
	\end{tabular}
	\vspace{-0.3cm}
\end{table}

\noindent \textbf{Task-Aware Behavior of Adaptive Computation.} To further investigate the dynamics of our macro-micro hierarchical framework, we quantitatively analyze the correlation between the macro-level early-exit probability ($\beta$) and specific task categories on the LongVideoBench. Following the task taxonomy introduced in the original benchmark \cite{wu2024longvideobench}, the adaptive mechanism exhibits distinct routing preferences depending on the semantic requirements of the query. For perception-centric tasks (Level-1) characterized by explicit temporal anchors—such as Text-Referred Event (T2E) and Object-Referred Event (O2E)—VideoMM triggers early exits at a consistently high rate (ranging from $76\%$ to $83\%$). Circumventing the fine-grained micro level in these localized tasks significantly reduces inference latency. This phenomenon suggests that coarse-grained proxy tokens are highly effective for extracting localized semantics, whereas enforcing full-resolution inference may inadvertently cause the model to over-assimilate irrelevant visual noise. Conversely, for relation-centric tasks (Level-2) that necessitate global temporal logic—such as Sequence of Scenes (SSS) and Object Before/After Object (O3O)—the early-exit rate decreases to approximately $60\%$. In these complex scenarios, the macro-level consensus naturally diverges. This divergence inherently stems from the fact that the spatial downsampling within the macro proxy often obscures subtle visual state transitions and fine-grained action boundaries essential for stringent temporal deduction. Recognizing this semantic deficit, the framework conservatively routes the samples to the micro level for comprehensive tracking, ensuring lossless visual features for long-range dependencies. Consequently, the macro-micro paradigm functions as an implicit task-aware router: it accelerates and filters noise for localized queries while prudently preserving high-resolution computational capacity for long-span reasoning. From a system optimization perspective, this dynamic routing profoundly regulates the inference memory footprint.

\begin{table}[t]
    \centering
    \caption{Quantitative analysis of early-exit behaviors across representative tasks on LongVideoBench. The macro-micro paradigm naturally exhibits higher early-exit rates for localized perception tasks, while conservatively maintaining deep inference (lowest $\beta$) for complex relation tasks.}
    \vspace{-0.1cm}
    \label{tab:task_early_exit}
    \resizebox{0.8\columnwidth}{!}{%
    \renewcommand{\arraystretch}{1.1}
    \setlength{\tabcolsep}{6pt}
    \begin{tabular}{@{}llc@{}}
        \toprule
        \textbf{Task Level} & \textbf{Task} & \textbf{Early-Exit Rate ($\beta$)} \\
        \midrule
        \multirow{3}{*}{\begin{tabular}[c]{@{}l@{}}Perception \\ (Level-1)\end{tabular}} 
        & T2O (Text-Obj) & $76.32\%$ \\
        & T2E (Text-Event) & $83.08\%$ \\
        & O2E (Obj-Event) & $82.76\%$ \\
        \midrule
        \multirow{3}{*}{\begin{tabular}[c]{@{}l@{}}Relation \\ (Level-2)\end{tabular}}   
        & T3O (Text-Obj) & $63.51\%$ \\
        & SSS (Scene Seq) & $62.89\%$ \\
        & O3O (Obj-Obj) & $60.61\%$ \\
        \bottomrule
    \end{tabular}%
    }
    \vspace{-0.3cm}
\end{table}

\section{Conclusion}

In this work, we identified a fundamental dilemma in efficient long-form video understanding: existing token reduction methods rely on model-centric downsizing for token selection, failing to reconcile the trade-off between selection precision and computational overhead.
To address this, we proposed \textbf{VideoMM}, a novel framework that orchestrates a paradigm shift from model scale to adaptive perceptual granularity. By adhering to a coarse-to-fine philosophy, VideoMM decouples semantic filtering from dense processing via a cost-effective \textit{Macro Proxy}, while the \textit{Adaptive Macro-Micro Inference} mechanism dynamically recruits high-fidelity tokens only when necessary.Extensive evaluations on LongVideoBench and other benchmarks demonstrate that VideoMM not only breaks the efficiency bottleneck—achieving a $6.13\times$ speedup—but also improves accuracy by over 7\% compared to full-context baselines. Furthermore, our in-depth analysis of multiple variants under the Macro-Micro design demonstrates that the VideoMM family establishes a superior Pareto frontier compared to prior arts, empirically confirming the inherent effectiveness of the \textit{hierarchical perceptual granularity paradigm}. We believe this promising perspective opens a new avenue for efficient video understanding,  facilitating the real-world deployment of multimodal systems in long-context scenarios.

\newpage

\bibliographystyle{ACM-Reference-Format}
\bibliography{example_paper}

\appendix

\newpage
\appendix

\section{Additional Related Works.} 
\label{apdx:related}
MLLMs typically encode visual inputs as discrete visual tokens. Early approaches were restricted to processing videos at fixed resolutions, such as InternVL \cite{InternVL} and LLaVA-NeXT \cite{liu2024llavanext}, which hinders the comprehension of content containing diverse semantic details. Most recent advancements support native dynamic resolution, enabling direct, patch-wise partitioning of inputs at arbitrary resolutions. Indeed, this capability has emerged as the prevailing paradigm in contemporary MLLMs. Leading open-source mllms, including Qwen3-VL \cite{qwen3technicalreport} and GLM-4V \cite{glm4v}, have adopted this paradigm, enabling direct patch-wise tokenization of visual inputs at arbitrary resolutions. This capability of mllms to handle videos at arbitrary native resolutions serves as the foundation for our VideoMM. Leveraging this flexibility, VideoMM adaptively modulates video resolution to coordinate video inference on a coarse-to-fine basis, enhancing comprehensive video understanding.

Beyond token reduction, a significant body of work addresses long-context LLM inference through complementary mechanisms that can be synergistically integrated with our approach.

\textbf{KV Cache Eviction} accelerates the decoding process by selectively pruning internally retained KV caches. Predominantly operating in long text scenarios, these methods optimize compression by assessing cache importance \cite{zhang2023h2oheavyhitteroracleefficient,li2024snapkv,feng2025identify,feng2025taming} or employing dynamic budget allocation~\cite{cai2024pyramidkv,feng2024ada,fu2024headkv}. Recent researches have been extended to MLLMs: notably, SparseMM \cite{wang2025sparsemm} incorporates inter-head allocation into multimodal cache management.
However, since these methods compress the cache only after processing all tokens to alleviate memory pressure, they do not reduce the token count itself. Consequently, they are orthogonal to token reduction strategies.

\textbf{Sparse attention methods} \cite{xiao2024infllm, tang2024quest, jiang2024minference, li2025mminferenceacceleratingprefillinglongcontext} accelerate inference by retaining the full KV cache while selecting only a critical subset for computation. However, as these approaches do not lower the number of processed tokens, they remain orthogonal to token reduction methods. Future work could combine lightweight token reduction with sparse attention to further minimize inference overhead.

\textbf{Speculative decoding} \cite{leviathan2023fastinferencetransformersspeculative, chen2023acceleratinglargelanguagemodel, zhang2024draft} accelerates inference by utilizing a small model to draft outputs, which are then verified by a larger model, thereby improving decoding efficiency through collaborative inference. This technique has recently been extended to MLLMs. For example, SpecVLM \cite{ji2025specvlm} enhances decoding efficiency by compressing visual tokens and passing them to a smaller model for drafting, followed by verification from a larger model. This demonstrates that visual token reduction is orthogonal to the speculative decoding paradigm. Future work could explore integrating these two approaches more closely.

\begin{table*}[t!]
	\centering
	\caption{Detailed performance comparison on LongVideoBench. The best two results are \textbf{bolded}.}
	\label{tab:longvideobench_details}
	\resizebox{0.99\textwidth}{!}{%
		\setlength{\tabcolsep}{4pt}
        \renewcommand{\arraystretch}{0.9}
		\begin{tabular}{@{}lcccccccccccccccccc@{}}
			\toprule
			\multirow{2}{*}{Method} & \multicolumn{18}{c}{LongVideoBench} \\
			\cmidrule(lr){2-19}
			 & E2O & E3E & O2E & O3O & S2A & S2E & S2O & SAA & SOS & SSS & T2A & T2E & T2O & T3E & T3O & TAA & TOS & Avg. \\
			\midrule
            \multicolumn{19}{c}{Qwen2.5-VL-7B}\\
            \midrule
            Vanilla & 67.69 & 56.38 & 62.07 & 53.03 & 67.05 & 66.67 & 61.11 & 56.94 & 66.67 & 40.21 & 62.03 & 64.62 & 55.26 & 50.68 & 52.70 & 58.54 & 39.73 & 57.67 \\
            \arrayrulecolor{lightgray}
            \midrule
            \arrayrulecolor{black}
            VisionZip & 73.44 & 61.96 & 63.22 & \textbf{60.61} & 67.05 & 68.82 & 51.39 & 52.78 & \textbf{69.14} & 43.16 & 60.76 & 64.62 & 52.63 & \textbf{49.32} & 56.76 & \textbf{54.32} & \textbf{39.73} & 58.23 \\
            VidCom & 61.54 & 61.70 & 59.77 & 48.48 & 64.77 & 70.97 & 52.78 & 51.39 & 61.73 & 36.08 & 58.23 & 66.15 & 51.32 & 41.10 & 48.65 & \textbf{57.32} & 32.88 & 54.60 \\
            Flexselect & \textbf{73.85} & \textbf{73.40} & \textbf{67.82} & \textbf{66.67} & \textbf{77.27} & \textbf{75.27} & \textbf{65.28} & \textbf{61.11} & \textbf{69.14} & \textbf{53.61} & \textbf{68.35} & \textbf{67.69} & \textbf{68.42} & 45.21 & \textbf{64.86} & 50.00 & 38.36 & \textbf{64.10} \\
            \rowcolor{gray!15}VideoMM & \textbf{76.92} & \textbf{73.40} & \textbf{68.97} & 56.06 & \textbf{79.55} & \textbf{73.12} & \textbf{65.28} & \textbf{63.89} & \textbf{69.14} & \textbf{54.64} & \textbf{70.89} & \textbf{67.69} & \textbf{68.42} & \textbf{52.05} & \textbf{59.46} & 53.66 & \textbf{42.47} & \textbf{64.70} \\
            \midrule
            \multicolumn{19}{c}{GLM-4.1V-9B}\\
            \midrule
            Vanilla & 58.46 & 67.02 & 72.09 & 57.58 & 75.00 & 62.37 & 55.56 & 56.94 & 64.20 & 38.14 & 67.09 & 61.54 & 55.26 & 50.68 & 55.41 & 62.20 & 39.73 & 58.98 \\
            \arrayrulecolor{lightgray}
            \midrule
            \arrayrulecolor{black}
            VisionZip & 53.23 & 61.54 & 62.07 & 52.38 & 66.27 & 61.11 & 54.29 & 49.30 & 67.50 & 37.89 & 51.32 & \textbf{60.32} & 51.39 & 53.52 & \textbf{55.41} & 62.50 & 38.03 & 55.35 \\
            VidCom & 55.38 & 60.64 & 56.32 & 45.45 & 60.23 & 61.29 & 44.44 & 51.39 & 58.02 & 32.99 & 55.70 & 53.85 & 50.00 & 45.21 & 47.30 & 57.32 & \textbf{39.73} & 51.68 \\
            Flexselect & \textbf{56.92} & \textbf{63.83} & \textbf{63.22} & \textbf{62.12} & \textbf{73.86} & \textbf{67.74} & \textbf{59.72} & \textbf{58.33} & \textbf{67.90} & \textbf{45.36} & \textbf{77.22} & 60.00 & \textbf{52.63} & \textbf{57.53} & \textbf{55.41} & \textbf{69.51} & 35.62 & \textbf{60.66} \\
            \rowcolor{gray!15}VideoMM & \textbf{69.23} & \textbf{74.47} & \textbf{71.26} & \textbf{60.61} & \textbf{80.68} & \textbf{72.04} & \textbf{62.50} & \textbf{55.56} & \textbf{72.84} & \textbf{46.39} & \textbf{72.15} & \textbf{67.69} & \textbf{60.53} & \textbf{54.79} & \textbf{59.46} & \textbf{69.51} & \textbf{38.36} & \textbf{64.32} \\
            \midrule
            \multicolumn{19}{c}{Qwen3-VL-8B}\\
            \midrule
            Vanilla & 72.31 & 78.72 & 72.41 & 60.61 & 82.95 & 68.82 & 72.22 & 65.28 & 70.37 & 57.73 & 74.68 & 66.15 & 71.05 & 53.42 & 60.81 & 59.76 & 45.21 & 66.94 \\
            \arrayrulecolor{lightgray}
            \midrule
            \arrayrulecolor{black}
            VisionZip & 59.68 & 71.91 & 67.06 & 58.06 & 71.25 & 65.17 & 55.71 & 58.33 & 64.10 & 49.47 & 58.67 & 63.49 & 63.89 & 42.03 & 51.43 & 57.50 & 39.44 & 58.97 \\
            VidCom & 67.69 & 65.96 & 64.37 & 56.06 & 70.45 & \textbf{70.97} & 61.11 & 54.17 & 64.20 & 43.30 & 64.56 & 64.62 & 65.79 & 43.84 & 56.76 & 51.22 & 38.36 & 59.16 \\
            Flexselect & \textbf{73.85} & \textbf{76.60} & \textbf{68.97} & \textbf{69.70} & \textbf{81.82} & \textbf{72.04} & \textbf{65.28} & \textbf{66.67} & \textbf{70.37} & \textbf{58.76} & \textbf{78.48} & \textbf{66.15} & \textbf{75.00} & \textbf{49.32} & \textbf{63.51} & \textbf{58.54} & \textbf{42.47} & \textbf{67.17} \\
            \rowcolor{gray!15}VideoMM & \textbf{70.77} & \textbf{77.66} & \textbf{72.41} & \textbf{68.18} & \textbf{80.68} & \textbf{70.97} & \textbf{69.44} & \textbf{63.89} & \textbf{71.60} & \textbf{54.64} & \textbf{75.95} & \textbf{73.85} & \textbf{73.68} & \textbf{47.95} & \textbf{64.86} & \textbf{62.20} & \textbf{46.58} & \textbf{67.54} \\
			\bottomrule
		\end{tabular}%
	}
\end{table*}

\begin{table*}[t!]
	\centering
	\caption{Detailed performance comparison on LVBench. The best two results are \textbf{bolded}.}
	\label{tab:lvbench_details}
	\resizebox{0.99\textwidth}{!}{%
		\setlength{\tabcolsep}{6pt}
        \renewcommand{\arraystretch}{0.9}
		\begin{tabular}{@{}lccccccc@{}}
			\toprule
			\multirow{2}{*}{Method} & \multicolumn{7}{c}{LVBench} \\
			\cmidrule(lr){2-8}
			 & Key Info. Retrieval & Event Understanding & Summarization & Entity Recognition & Reasoning & Temporal Grounding & Avg. \\
			\midrule
            \multicolumn{8}{c}{Qwen2.5-VL-7B}\\
            \midrule
            Vanilla & 45.02 & 36.79 & 32.76 & 39.44 & 39.80 & 30.91 & 39.57 \\
            \arrayrulecolor{lightgray}
            \midrule
            \arrayrulecolor{black}
            VisionZip & 41.92 & 38.79 & 32.76 & 41.95 & 42.29 & 36.82 & 41.38 \\
            VidCom & 47.08 & 39.10 & 36.21 & 39.14 & 36.32 & 36.82 & 40.80 \\
            Flexselect & \textbf{58.76} & \textbf{47.14} & \textbf{44.83} & \textbf{51.99} & \textbf{47.26} & \textbf{39.55} & \textbf{51.45} \\
            \rowcolor{gray!15}VideoMM & \textbf{58.42} & \textbf{47.45} & \textbf{39.66} & \textbf{51.40} & \textbf{49.75} & \textbf{41.36} & \textbf{50.94} \\
            \midrule
            \multicolumn{8}{c}{GLM-4.1V-9B}\\
            \midrule
            Vanilla & 52.58 & 45.75 & 39.66 & 47.71 & 47.76 & 36.82 & 47.90 \\
            \arrayrulecolor{lightgray}
            \midrule
            \arrayrulecolor{black}
            VisionZip & 35.05 & 35.39 & \textbf{31.03} & 37.08 & 37.31 & 27.73 & 36.41 \\
            VidCom & 46.05 & 36.32 & \textbf{32.76} & 34.12 & 43.28 & 39.55 & 37.70 \\
            Flexselect & \textbf{55.33} & \textbf{48.69} & \textbf{31.03} & \textbf{50.66} & \textbf{48.76} & \textbf{41.82} & \textbf{50.16} \\
            \rowcolor{gray!15}VideoMM & \textbf{53.95} & \textbf{48.22} & 29.31 & \textbf{49.78} & \textbf{49.75} & \textbf{40.91} & \textbf{49.52} \\
            \midrule
            \multicolumn{8}{c}{Qwen3-VL-8B}\\
            \midrule
            Vanilla & 60.82 & 51.47 & 32.76 & 56.72 & 44.78 & 47.73 & 54.10 \\
            \arrayrulecolor{lightgray}
            \midrule
            \arrayrulecolor{black}
            VisionZip & 39.52 & 41.27 & 27.59 & 41.06 & 37.31 & 35.45 & 40.22 \\
            VidCom & 52.23 & 43.28 & 29.31 & 44.02 & 39.30 & 40.00 & 44.03 \\
            Flexselect & \textbf{63.23} & \textbf{52.55} & \textbf{34.48} & \textbf{56.28} & \textbf{52.74} & \textbf{46.82} & \textbf{55.20} \\
            \rowcolor{gray!15}VideoMM & \textbf{58.42} & \textbf{52.86} & \textbf{31.03} & \textbf{57.16} & \textbf{48.26} & \textbf{45.45} & \textbf{54.36} \\
			\bottomrule
		\end{tabular}%
	}
\end{table*}

\begin{table*}[t!]
	\centering
	\caption{Detailed performance comparison on VideoMME. The best two results are \textbf{bolded}.}
	\label{tab:videomme_details}
	\resizebox{0.99\textwidth}{!}{%
		\setlength{\tabcolsep}{4pt}
        \renewcommand{\arraystretch}{0.9}
		\begin{tabular}{@{}lccccccccccccc@{}}
			\toprule
			\multirow{2}{*}{Method} & \multicolumn{13}{c}{VideoMME} \\
			\cmidrule(lr){2-14}
			 & Counting Problem & Info. Synopsis & Object Recog. & Action Reason. & Object Reason. & Temporal Percep. & Attribute Percep. & Temporal Reason. & Action Recog. & OCR Problems & Spatial Percep. & Spatial Reason. & Avg. \\
			\midrule
            \multicolumn{14}{c}{Qwen2.5-VL-7B}\\
            \midrule
            Vanilla & 40.67 & 79.26 & 68.36 & 53.33 & 56.83 & 69.09 & 72.52 & 41.24 & 63.58 & 68.35 & 62.96 & 80.36 & 61.56 \\
            \arrayrulecolor{lightgray}
            \midrule
            \arrayrulecolor{black}
            VisionZip & 37.31 & 78.02 & 70.06 & 55.44 & 57.49 & \textbf{80.00} & 73.87 & 50.28 & 62.62 & 64.75 & 59.26 & 75.00 & 62.07 \\
            VidCom & 38.06 & 74.30 & 67.80 & 56.14 & 58.81 & \textbf{74.55} & 67.57 & 49.15 & 65.18 & 71.22 & 61.11 & 69.64 & 61.56 \\
            Flexselect & \textbf{45.90} & \textbf{81.11} & \textbf{74.58} & \textbf{63.51} & \textbf{65.86} & \textbf{74.55} & \textbf{79.73} & \textbf{61.02} & \textbf{68.37} & \textbf{79.86} & \textbf{64.81} & \textbf{78.57} & \textbf{68.85} \\
            \rowcolor{gray!15}VideoMM & \textbf{46.64} & \textbf{81.42} & \textbf{72.32} & \textbf{59.65} & \textbf{66.74} & 72.73 & \textbf{77.93} & \textbf{60.45} & \textbf{69.97} & \textbf{79.14} & \textbf{66.67} & \textbf{80.36} & \textbf{68.41} \\
            \midrule
            \multicolumn{14}{c}{GLM-4.1V-9B}\\
            \midrule
            Vanilla & 44.78 & 79.57 & 75.14 & 55.79 & 63.22 & 70.91 & 77.48 & 54.24 & 67.09 & 80.58 & 68.52 & 80.36 & 66.67 \\
            \arrayrulecolor{lightgray}
            \midrule
            \arrayrulecolor{black}
            VisionZip & 35.07 & 73.07 & 66.10 & \textbf{55.09} & 55.51 & \textbf{65.45} & 73.42 & 45.20 & 62.30 & 58.99 & 64.81 & 75.00 & 59.48 \\
            VidCom & 35.07 & 72.45 & 56.78 & 47.37 & 57.27 & 56.36 & 63.96 & 47.46 & 56.23 & 73.38 & 55.56 & 64.29 & 56.48 \\
            Flexselect & \textbf{43.28} & \textbf{77.09} & \textbf{70.90} & 50.88 & \textbf{64.32} & 63.64 & \textbf{75.68} & \textbf{54.24} & \textbf{67.73} & \textbf{80.58} & \textbf{66.67} & \textbf{83.93} & \textbf{65.15} \\
            \rowcolor{gray!15}VideoMM & \textbf{42.54} & \textbf{79.26} & \textbf{73.73} & \textbf{59.65} & \textbf{64.32} & \textbf{70.91} & \textbf{78.38} & \textbf{57.06} & \textbf{68.05} & \textbf{82.01} & \textbf{70.37} & \textbf{83.93} & \textbf{67.37} \\
            \midrule
            \multicolumn{14}{c}{Qwen3-VL-8B}\\
            \midrule
            Vanilla & 52.61 & 83.59 & 76.84 & 63.16 & 69.38 & 81.82 & 81.08 & 63.84 & 70.61 & 82.73 & 72.22 & 83.93 & 71.78 \\
            \arrayrulecolor{lightgray}
            \midrule
            \arrayrulecolor{black}
            VisionZip & 41.04 & 79.57 & 69.21 & 60.35 & 63.66 & 74.55 & 80.18 & 54.24 & 66.45 & 64.03 & 70.37 & 80.36 & 65.48 \\
            VidCom & 45.52 & 78.33 & 70.62 & 56.84 & 63.00 & 70.91 & 73.42 & 56.50 & 63.58 & 66.19 & 68.52 & 73.21 & 64.59 \\
            Flexselect & \textbf{50.75} & \textbf{84.21} & \textbf{79.38} & \textbf{63.16} & \textbf{69.60} & \textbf{76.36} & \textbf{81.98} & \textbf{67.80} & \textbf{69.65} & \textbf{82.73} & \textbf{75.93} & \textbf{82.14} & \textbf{72.19} \\
            \rowcolor{gray!15}VideoMM & \textbf{50.37} & \textbf{83.59} & \textbf{75.42} & \textbf{63.51} & \textbf{69.38} & \textbf{83.64} & \textbf{82.43} & \textbf{65.54} & \textbf{71.88} & \textbf{80.58} & \textbf{72.22} & \textbf{82.14} & \textbf{71.67} \\
			\bottomrule
		\end{tabular}%
	}
\end{table*}

\textbf{Agent-based Video Understanding} \cite{chen2025lvagent, wang2026video, wang2026videohv, yan2026symphony} tackles long-video redundancy at a semantic level through interactive reasoning. LVAgent \cite{chen2025lvagent} introduces a multi-round dynamic collaboration framework among multiple MLLM agents, filtering out suboptimal reasoning through iterative selection, perception, action, and reflection. VideoChat-A1 \cite{wang2026video} emphasizes the inherent shot-based structure of videos by proposing a Chain-of-Shot reasoning paradigm. It progressively selects relevant shots, partitions them into fine-grained subshots via feature clustering, and evaluates reasoning confidence to iteratively refine the temporal context. VideoHV-Agent \cite{wang2026videohv} reformulates long-video question answering as a hypothesis--verification process: a Thinker converts candidate answers into testable hypotheses, a Judge derives discriminative clues, and a Verifier localizes and examines targeted video evidence before an Answer agent integrates the results. Symphony \cite{yan2026symphony} employs a cognitively inspired multi-agent architecture that coordinates task decomposition, evidence grounding, visual perception, and reflection for long-video reasoning. Future work could explore synergizing agent-based semantic exploration with visual token reduction methods for optimal long-video inference efficiency.

\section{Benchmark Details.} 
Below, we provide a detailed overview of the video understanding benchmarks utilized in the experiments.
\begin{enumerate}
\item \textbf{LongVideoBench \cite{wu2024longvideobench}} is a comprehensive video question-answering benchmark comprising 3,763 web-collected videos (up to one hour) with 6,678 human-annotated multiple-choice questions across 17 fine-grained categories. It introduces a referring reasoning task where models must retrieve and reason over relevant multimodal information from specified video contexts.

\item \textbf{VideoMME \cite{fu2025video}} is a full-spectrum multi-modal evaluation benchmark for MLLMs comprising 900 videos (254 hours) with 2,700 question-answer pairs. It spans 6 visual domains, covers short- to long-form videos (11 seconds to 1 hour), and includes multi-modal inputs (frames, subtitles, audio).

\item \textbf{LVBench \cite{wang2024lvbench}} is a benchmark specifically designed for long video understanding, created to address the gap in evaluating MLLMs for real-world applications requiring comprehension of videos spanning several hours. It comprises publicly sourced videos and includes diverse tasks focused on long video comprehension and information extraction.
\end{enumerate}

\section{Model Details}
Our evaluation employs three models, all of which support dynamic resolution:

\begin{itemize}
    \item \textbf{Qwen2.5-VL-7B-Instruct \cite{Qwen2.5-VL}} builds upon the Qwen2-VL architecture, introducing significant advancements in visual-language understanding. Key enhancements include superior comprehension of complex visual elements such as text, charts, and layouts, alongside agentic capabilities for dynamic tool direction. Architecturally, it extends dynamic resolution to the temporal dimension through dynamic Frame Per Second sampling, complemented by mRoPE updates that facilitate learning of temporal sequences and precise event identification within videos.
    
    \item \textbf{GLM4.1V-9B-Thinking \cite{glm4v}}: is a powerful Mllm designed for advanced general-purpose multimodal reasoning. At its core is a reasoning-centric training framework that leverages a robust vision foundation model, initially established through large-scale pre-training. This foundation's full potential is subsequently unlocked via Reinforcement Learning, significantly enhancing the model's comprehensive capabilities. It also supports dynamic-resolution video inputs.

    \item \textbf{Qwen3-VL-8B-Instruct \cite{qwen3technicalreport}} stands as the most powerful lite mllm in the Qwen series to date, featuring comprehensive upgrades. It incorporates a novel Interleaved-MRoPE, which provides full-frequency allocation across time, width, and height through robust positional embeddings, thereby enhancing long-horizon video reasoning. Additionally, it employs Text–Timestamp Alignment, moving beyond T-RoPE to achieve precise, timestamp-grounded event localization for stronger video temporal modeling. 
\end{itemize}

\begin{table}[t]
	\centering
	\caption{Impact of Lite Models on LongVideoBench based on Qwen3-VL-8B. Acc.: Accuracy (\%); Thpt.: Throughput (samples/min))}
	\vspace{0.2cm}
	\label{tab:lite_model}
	\small
	\setlength{\tabcolsep}{3pt}
	\renewcommand{\arraystretch}{1.0}
	\begin{tabular}{@{}lcc@{}}
		\toprule
		Method & Accuracy & Throughput \\
		\midrule
		Vanilla          & 66.94 & 0.85 \\
         \arrayrulecolor{lightgray}
        \midrule
        \arrayrulecolor{black}
		VideoMM         & 67.54 & 3.69 \\
		\bottomrule
	\end{tabular}
\end{table}

\section{Accelerating Token Compression Using Lightweight MLLMs} 
\label{apdx:var1}
While VideoMM primarily focuses on optimizing representation granularity, its framework is orthogonal to—and can be effectively combined with—model-centric downsizing strategies. In such a configuration, a lightweight model with fewer parameters conducts the initial token selection, streamlining the process before a larger MLLM performs the final inference. We try a direct integration. To demonstrate this flexibility, we instantiate a variant named \textbf{VideoMM-Lite}, where we directly employ the off-the-shelf Qwen3-VL-4B for token selection and the Qwen3-VL-8B for the final inference, avoiding additional training overhead \footnote{Although the distillation-based specialization strategy employed by FlexSelect~\cite{zhang2025flexselectflexibletokenselection} is equally applicable to our framework, we leave such optimization for future work to avoid additional training overhead, maintaining our primary focus on the efficacy of data granularity paradigm.}. As shown in Table~\ref{tab:lite_model}, compared to the standard VideoMM, VideoMM-Lite increases throughput from 3.69 to 4.19 samples/mins, with only a marginal accuracy dip from 67.54\% to 66.34\%. This result confirms that integrating a smaller parameter model for selection is a viable pathway to further enhance throughput with minimal impact on performance.

\begin{table}[t]
	\centering
	\caption{Impact of multi-stages of VideoMM on LongVideoBench.}
	\vspace{0.1cm}
	\label{tab:three_stage}
	\small
	\setlength{\tabcolsep}{3pt}
	\renewcommand{\arraystretch}{1.0}
	\begin{tabular}{@{}lcc@{}}
		\toprule
		Method & Accuracy & Throughput \\
		\midrule
		Vanilla          & 66.94 & 0.85 \\
		VideoMM (Two Stage)        & 67.54 & 3.69 \\
		VideoMM-multi (Three Stage)   & 64.55 & 5.91 \\ 
		\bottomrule
	\end{tabular}
	\vspace{0.2cm}
	\caption*{\small Accuracy represents the percentage of correct predictions, while Throughout denotes the number of samples processed per second.}
\end{table}

\begin{table}[t]
	\centering
	\caption{Comparison of VideoMM and FlexSelect on VideoMME using Qwen-VL models}
	\vspace{-0.1cm}
	\label{tab:comparison_fs_videomme}
    \footnotesize
	\setlength{\tabcolsep}{4pt}
	\renewcommand{\arraystretch}{1.0}
	\begin{tabular}{@{}lcc@{}}
		\toprule
		Method & Acc. & Thpt. \\
		\midrule

		\multicolumn{3}{c}{\textit{Qwen2.5-VL-7B}}\\
		\midrule
        Vanilla (512) & 61.56 & 0.85 \\
        FlexSelect (256) & 68.26 & 3.54 \\
        VideoMM (512) & 68.41 & 5.47 \\

        \midrule
		\multicolumn{3}{c}{\textit{GLM-4.1V-9B}}\\
		\midrule
        Vanilla (512) & 66.67 & 0.58 \\
        FlexSelect (256) & 69.56 & 3.25 \\
        VideoMM (512) & 67.37 & 4.21 \\

		\midrule
		\multicolumn{3}{c}{\textit{Qwen3-VL-8B}}\\
		\midrule
        Vanilla (512) & 71.78 & 0.97 \\
        FlexSelect (256) & 71.48 & 3.54 \\
        VideoMM (512) & 71.67 & 3.59 \\

		\bottomrule
	\end{tabular}
	\caption*{\small Acc.: Accuracy (\%); Thpt.: Throughput (samples/min)}
\end{table}

\section{Three-Stage Micro-Micro Exploration.} 
\label{apdx:var2}
To further demonstrate the extensibility of our framework, we introduce \textbf{VideoMM-multi}, a variant that expands the original two-stage design into a three-stage cascade. This is achieved by inserting an additional, coarser-grained hierarchical level prior to the standard Macro Proxy. Specifically, we implement an \textit{Ultra-Macro Proxy} with an aggressive downsampling factor of $k^2$ (compared to the standard factor $k$), designed to filter out the most obvious non-essential regions with extreme efficiency before passing the remaining candidates to the Macro level.We evaluated this three-stage design ($k=2$) using Qwen3-VL-8B on LongVideoBench. As shown in Table~\ref{tab:three_stage}, the comparison reveals a distinct trade-off between granular precision and computational speed. The standard two-stage VideoMM maintains superior semantic retention, achieving a higher accuracy of 67.54 compared to VideoMM-multi's 64.55. Conversely, by offloading the majority of  workload to the new Ultra-Macro tier, VideoMM-multi drastically reduces the computational burden, increasing throughput to 5.91 samples/min—significantly outperforming the standard VideoMM's 3.69 samples/min. These results confirm that dynamically adjusting the depth of the macro-micro hierarchy serves as a powerful lever for tuning the accuracy-efficiency trade-off.

\end{document}